%% file: paper.tex
\documentclass[10pt,a4paper,logo]{paper}

\usepackage[all]{hypcap}

\usepackage[authoryear, round]{natbib}

\usepackage{nicefrac}       % compact symbols for 1/2, etc.
\usepackage{multirow}
\usepackage{cleveref}
\usepackage{dsfont}
\usepackage{makecell}
\usepackage{subcaption}
\usepackage{bbm}
\usepackage{wrapfig}
\usepackage{mathtools,xparse}
\usepackage{array}
\usepackage{arydshln}
\usepackage{scalerel}
\usepackage{tablefootnote}
\usepackage{tocloft}
\usepackage{adjustbox}
\usepackage{setspace}
\usepackage[most,skins,theorems]{tcolorbox}

\usepackage{enumitem}
\usepackage{graphicx}
\usepackage{booktabs}
\usepackage{array}
\usepackage{microtype}
\usepackage[utf8]{inputenc}
\usepackage[T1]{fontenc}
\usepackage{subcaption}
\usepackage{amsmath} 
\usepackage{multirow}
\usepackage{inconsolata}
\usepackage{pifont}
\usepackage[utf8]{inputenc}
\usepackage{booktabs} % For professional lines
\usepackage{anyfontsize}

\newtcolorbox{promptbox}{
  colback=gray!5,
  colframe=gray!70,
  boxrule=0.8pt,
  arc=2mm,
  left=10pt,
  right=10pt,
  top=10pt,
  bottom=10pt,
  width=\linewidth,
  fontupper=\ttfamily\footnotesize,
  enhanced,
}

\newboolean{showsection}
\setboolean{showsection}{true}
\makeatletter
\@namedef{ver@everyshi.sty}{}
\makeatother

\definecolor{custom_green}{rgb}{0.0, 0.5, 0.0}
\definecolor{custom_red}{rgb}{1.0, 0.01, 0.24}
\definecolor{custom_blue}{HTML}{C9DAF7}
\definecolor{custom_purple}{HTML}{D9D1E9}
\definecolor{title_blue}{HTML}{204899}
\definecolor{cite_blue}{HTML}{044dc1}
\definecolor{cite_purple}{HTML}{7406a7}

\hypersetup{
    colorlinks = true,
    citecolor = {cite_blue},
    linkcolor = {cite_purple},
    urlcolor = {cite_purple},
}

\input{macro}

\input{math_commands.tex}

\makeatletter
\def\mathcolor#1#{\@mathcolor{#1}}
\def\@mathcolor#1#2#3{%
  \protect\leavevmode
  \begingroup
    \color#1{#2}#3%
  \endgroup
}
\makeatother

\Crefformat{equation}{#2Eq.\;(#1)#3}
\Crefformat{figure}{#2Figure #1#3}
\Crefformat{assumption}{#2Assumption #1#3}
\Crefname{assumption}{Assumption}{Assumptions}

\usepackage{crossreftools}
\makeatletter
\renewcommand\footnoterule{%
  \kern 15\p@
  \hrule \@width 2in \kern 2.6\p@
  \vspace{4pt}
}
\makeatother

\title{SentZero: An Enhanced Sentence-Centric Vision-Language Pretraining for Multi-Task Zero-Shot Chest X-Ray Analysis}

\reportnumber{}

\author[1]{Hangyul Yoon}
\author[1]{Hyungyung Lee}
\author[1]{Edward Choi}
\author[1,\footnotesize{$\dagger$}]{Eunho Yang}

\affil[1]{Kim Jaechul Graduate School of AI, Korea Advanced Institute of Science and Technology (KAIST)}

\input{sections/1_abstract}

\begin{document}

\maketitle

\vspace{-0.3cm}
\input{sections/2_introduction}
\input{sections/3_related_work}
\input{sections/4_method}
\input{sections/5_experiments}
\input{sections/6_conclusion}

\bibliographystyle{plainnat}
\bibliography{paper}

\clearpage
\appendix{\input{sections/7_appendix}}

\end{document}

%% file: macro.tex
\definecolor{blanchedalmond}{rgb}{1.0, 0.92, 0.8}
\definecolor{carmine}{rgb}{0.59, 0.0, 0.09}
\definecolor{lightblue}{rgb}{0.22,0.45,0.70}%

\renewcommand{\mathbf}{\boldsymbol}

\makeatletter
\def\Ddots{\mathinner{\mkern1mu\raise\p@
\vbox{\kern7\p@\hbox{.}}\mkern2mu
\raise4\p@\hbox{.}\mkern2mu\raise7\p@\hbox{.}\mkern1mu}}
\makeatother

\numberwithin{equation}{section}

\definecolor{amaranth}{rgb}{0.9, 0.17, 0.31}
\definecolor{antiquebrass}{rgb}{0.8, 0.58, 0.46}
\definecolor{antiquefuchsia}{rgb}{0.57, 0.36, 0.51}
\definecolor{chromeyellow}{rgb}{0.31, 0.47, 0.26}

\newcommand{\1}{\mathds 1}

%% file: math_commands.tex
\usepackage{amsmath,amsfonts,bm}

\def\eqref#1{equation~\ref{#1}}
\def\1{\bm{1}}

\DeclareMathAlphabet{\mathsfit}{\encodingdefault}{\sfdefault}{m}{sl}
\SetMathAlphabet{\mathsfit}{bold}{\encodingdefault}{\sfdefault}{bx}{n}

%% file: sections/1_abstract.tex
\begin{abstract}
Vision–language (VL) pretraining using paired chest X-ray (CXR) images and radiology reports has shown strong potential for medical image understanding. However, existing methods often remain dependent on task-specific fine-tuning because radiology reports are lengthy, clinically dense, and difficult to align with simple zero-shot prompts. Recent sentence-level approaches partially address this limitation using clinical phrases extracted by large language models (LLMs), but they largely overlook the intrinsic characteristics of radiology discourse. In particular, limited positive-pair diversity constrains further gains, while clinically equivalent sentences frequently recur across patients, creating false negatives in contrastive learning. To address these issues, we propose \textbf{SentZero}, an enhanced sentence-centric VL pretraining framework for zero-shot, multi-task CXR analysis. SentZero introduces LLM-based abstract-level sentence structuring and mapping to expand positive-pair diversity, together with an additional loss term to mitigate false negatives. We further introduce sentence-conditioned residual modulation of visual embeddings, enabling visual features to adapt to the semantic characteristics of each input sentence. Across diverse downstream tasks and datasets, SentZero improves zero-shot generalization and outperforms prior multi-task zero-shot methods.
\end{abstract}

%% file: sections/2_introduction.tex
\section{Introduction}
\label{sec:intro}
\vspace{-0.1cm}
Chest X-ray (CXR) remains one of the most widely used imaging modalities in clinical practice. With the rapid progress of vision--language (VL) pretraining, recent studies have leveraged CXR images paired with their corresponding captions, namely \textit{radiology reports} written by expert radiologists~\citep{zhang2022contrastive, wang2022multi, cheng2023prior, li2024mlip, liu2024towards, zhang2025medunifier}. By learning contrastive alignments between images and textual descriptions, these approaches have improved visual representations and benefited various downstream tasks. However, many existing methods still require fine-tuning on labeled datasets for specific applications~\citep{cheng2023prior, zhang2025medunifier, liu2024towards}, even after the contrastive pretraining. A key reason is that radiology reports are long and complex, containing detailed findings, clinical context, and stylistic variations across reporters. As a result, image–text alignment in the medical domain is noisier and more difficult than in general-domain VL pretraining, where concise prompts such as "There is a dog" can be used directly for zero-shot recognition. Consequently, applying simple text prompts to CXR tasks in a zero-shot manner remains challenging, often requiring additional task-specific fine-tuning. This reliance partially undermines the fundamental goal of VL pretraining: enabling broad generalization without task-specific supervision.

\vspace{-0.1cm}
To address this limitation, recent studies have explored zero-shot multi-task VL frameworks, in which a single model generalizes across multiple visual tasks using natural language prompts~\citep{huang2021gloria, zhang2022contrastive, zhang2023knowledge, wu2023medklip}. A pivotal milestone in this direction is the use of large language models (LLMs) for report rephrasing~\citep{lai2024carzero, park2025radzero}. By standardizing radiological findings into unified prompt templates, LLM-based rephrasing reduces noise from lengthy reports and diverse expressions, extracting multiple sentences from a single report. With sentence-level supervision, this design improves interpretability and enables zero-shot grounding within a unified framework, establishing the current state of the art.

\vspace{-0.1cm}
Despite these advances, existing approaches largely overlook a distinctive property of radiology discourse. Unlike general-domain captions, radiology reports are highly task-oriented and describe a narrow, well-defined set of findings, so identical or clinically equivalent statements, such as \textit{``The lungs are clear,''} recur across many patients. Current LLM-phrase-based methods treat each extracted sentence as an independent caption of its own image, which leads to two problems. First, the hierarchical semantic structure of radiology reports remains unexploited as a source of supervision. Second, clinically equivalent sentences from different studies are treated as negatives, creating \textit{false-negative pairs} that are wrongly pushed apart during contrastive training. Because the space of distinct clinical descriptions is inherently narrow, such pairs are frequent and cannot be avoided simply by better phrase extraction.

\vspace{-0.1cm}
We introduce \textbf{SentZero}, a \textit{sentence-centric} VL pretraining framework that uses semantic redundancy as a source of structured supervision. First, \textit{abstract-level sentence mapping} uses an LLM to map detailed phrases to concise topic--presence statements (e.g., \textit{`There is mild opacity in the bilateral lung base''} $\rightarrow$ \textit{`There is opacity''}). This provides supervision at two levels of granularity—detailed findings from the original text and their underlying clinical concepts—and enables matching of shared statements across studies. Second, we use these matches for false-negative mitigation through an auxiliary loss that selectively attracts the most relevant patches toward each shared statement. Rather than masking or relabeling such pairs at the instance level, which we find disrupts contrastive training, this auxiliary loss mitigates the effects of repeated clinical statements through selective patch-level alignment while leaving the global contrastive objective intact. Finally, \textit{sentence-conditioned residual modulation} applies sentence-dependent scale and shift parameters to attention-pooled visual features, allowing each sentence to guide \textit{how} the attended features are represented.

In summary, our contributions are as follows:
\vspace{-0.2cm}
\begin{itemize}[leftmargin=1.5em]
\item We propose SentZero, a sentence-centric VL pretraining framework for zero-shot, multi-task CXR analysis. Unlike most existing VL pretraining methods in the CXR domain, SentZero transfers directly to a variety of downstream tasks without any task-specific fine-tuning.
\item We leverage the shared semantic structure of radiology reports through abstract-level sentence mapping and a patch-level false-negative mitigation objective, complemented by sentence-conditioned residual modulation of visual features. We show that directly masking or relabeling false-negative pairs can impair performance, whereas our proposed selective patch-level attraction improves performance while preserving the original contrastive objective and pair assignments.
\item SentZero outperforms prior zero-shot multi-task methods on most evaluated benchmarks. Given the persistent challenge of strong zero-shot generalization across CXR tasks, these results mark a step toward a general-purpose CXR encoder that supports diverse tasks without additional task-specific annotations.
\end{itemize}

%% file: sections/3_related_work.tex
\section{Related Works}
\label{sec:related_work}
\paragraph{Vision-Language Pretraining in Chest X-ray.} In recent years, increasing attention has been given to leveraging paired image--text data in CXRs. In this setting, the textual modality consists of radiology reports, which are structured clinical descriptions routinely written by radiologists. With the release of large-scale paired datasets such as MIMIC-CXR~\citep{mimic_cxr_physionet, johnson2019mimic}, numerous CLIP-style~\citep{radford2021learning} VL pretraining methods have been developed for the medical domain~\citep{zhang2022contrastive, wang2022multi, cheng2023prior, li2024mlip, liu2024towards, zhang2025medunifier}. These studies consistently show that visual encoders pretrained through image--report alignment adapt better to downstream medical tasks than those pretrained on general-domain datasets such as ImageNet.

Despite these achievements, most existing VL pretraining methods still require additional task-specific supervised fine-tuning. One key reason is that radiology reports differ substantially from general-domain image captions: they are written for diagnosis and rigorous assessment of patient status, making them lengthy, detailed, and clinically dense. Such complexity can introduce noise during VL pretraining and makes it difficult to directly apply simple text prompts for zero-shot CXR tasks. As a result, many prior studies have relied on labeled downstream datasets to fine-tune pretrained models, partially undermining the original goal of VL pretraining: learning broadly generalizable representations with minimal task-specific supervision.

\vspace{-0.3cm}
\paragraph{Chest X-ray Pretraining for Zero-Shot Transfer.}
To reduce reliance on task-specific supervised fine-tuning after VL pretraining, several studies have explored zero-shot generalization across diverse CXR analysis tasks. Early methods, including MedKLIP~\citep{wu2023medklip} and KAD~\citep{zhang2023knowledge}, incorporated medical knowledge and clinical entities to improve semantic alignment. CARZero~\citep{lai2024carzero} introduced cross-attention-based alignment and LLM-driven phrase extraction, standardizing heterogeneous diagnostic expressions into sentence-level prompts for zero-shot classification and grounding. RadZero~\citep{park2025radzero} extended this paradigm through similarity-weighted patch aggregation and multi-positive contrastive learning~\citep{lee2022uniclip}, aligning each image with multiple finding sentences.

A key challenge in sentence-level pretraining is that clinically equivalent statements recur across patients, causing valid image--sentence associations to be treated as negatives. CoNNs~\citep{lian2026concept} addresses this issue using structured clinical concepts to relabel or exclude cross-patient pairs from the contrastive objective. However, we empirically show that directly relabeling or masking such pairs can instead degrade performance (Sec.~\ref{sec:result_false_neg_comparison}). SentZero explores a complementary approach: retaining the contrastive pair assignments while applying an auxiliary attraction loss to the highest-similarity patches of pairs identified through shared sentences. Furthermore, SentZero extends text conditioning beyond spatial attention by applying sentence-conditioned residual modulation to the aggregated visual features before contrastive comparison. These mechanisms provide local alignment supervision for shared clinical statements and sentence-dependent adaptation of visual representations.

\begin{figure*}[t]
    \centering
    \includegraphics[width=\textwidth]{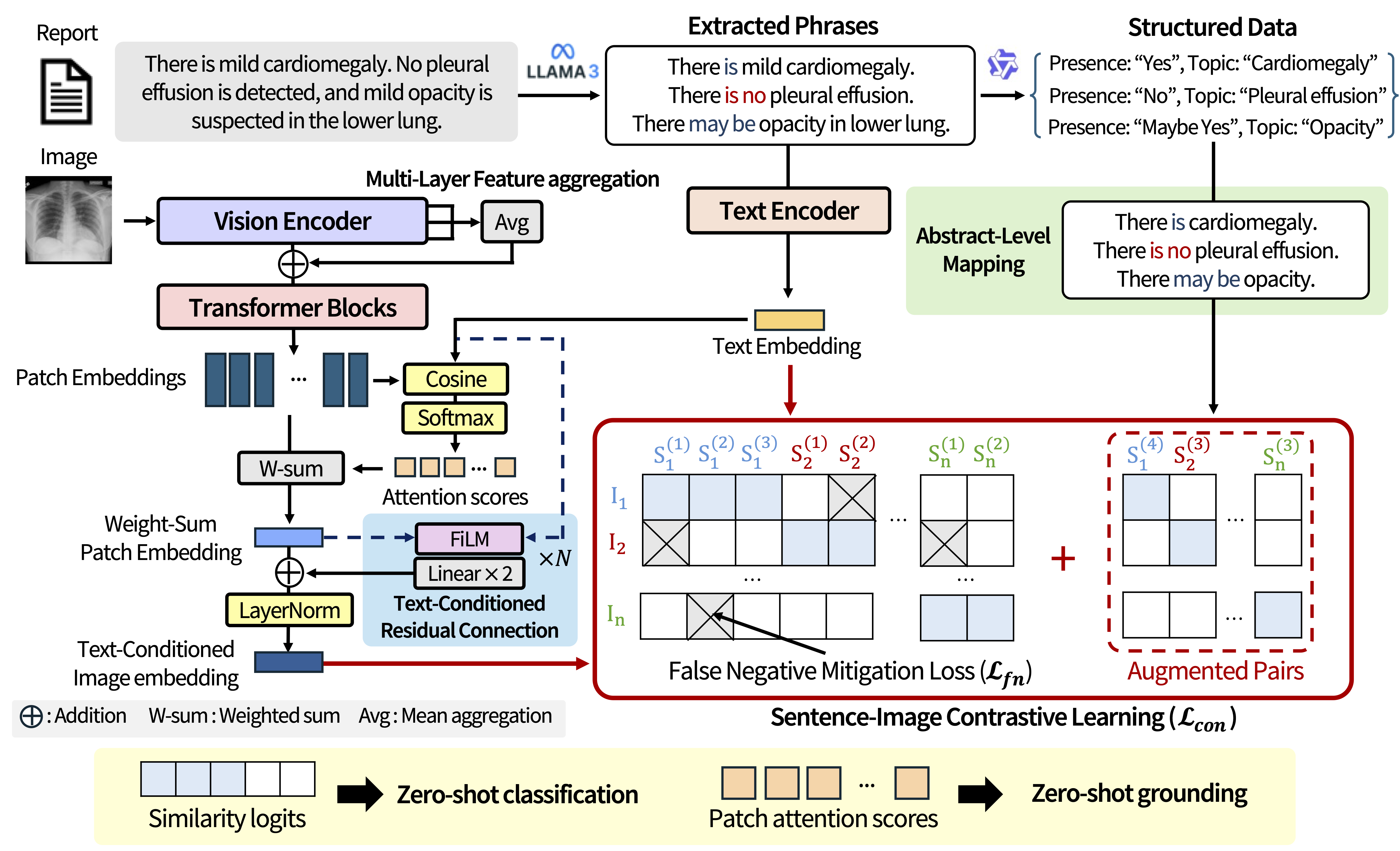}
    \vspace{-0.5cm}
    \caption{Overview of the SentZero training framework. SentZero performs multi-pair contrastive learning between sentence embeddings and weighted-sum patch embeddings (bold red line, lower-right panel). Structured tuples extracted from the findings generate augmented positive pairs through abstract-level mapping. The model is trained with a combination of the contrastive loss ($\mathcal{L}_{con}$) and the false-negative mitigation loss ($\mathcal{L}_{fn}$). Before contrastive learning, the patch embeddings are refined via a residual connection with a text-conditioning module (lower left). At inference, the similarity logits and patch attention scores support multiple zero-shot downstream tasks.}
    \label{fig:main}
    \vspace{-0.55cm}
\end{figure*}

%% file: sections/4_method.tex
\section{Method}
\label{sec:method}
The overall training framework is illustrated in Fig.~\ref{fig:main}. Given lengthy and noisy radiology reports, we first use an LLM to extract phrase-level clinical expressions (Sec.~\ref{sec:method_phrase_extraction}). We then employ the LLM to structure the extracted information and perform abstract-level sentence mapping, thereby augmenting positive pairs while preserving the implicit medical semantics of the sentences. The resulting sentence and image embeddings (Sec.~\ref{sec:method_feature_extraction}) are used for multi-pair contrastive learning, where the image embeddings are conditioned to reflect the semantics of each sentence rather than relying on the original patch embeddings (Sec.~\ref{sec:loss_calculation}). During training, false-negative pairs are identified and calibrated using a false-negative mitigation loss, addressing an issue that has not been adequately handled in previous studies (Sec.~\ref{sec:false_neg_mitigation}). Once trained, the model can be directly applied to zero-shot classification and spatial grounding tasks without task-specific fine-tuning (Sec.~\ref{sec:method_inference}).

\vspace{-0.2cm}
\subsection{Abstract-Level Sentence Mapping}
\label{sec:method_phrase_extraction}
\vspace{-0.1cm}
Let $\mathcal{D}_{\text{batch}} = \{(I_i, R_i)\}_{i=1}^B$ denote a mini-batch of $B$ paired images and radiology reports. For each report $R_i$ corresponding to the $i$-th image $I_i$, we employ an LLM to extract a set of $M_i$ phrases, defined as $T_i = \{S_i^{(1)}, S_i^{(2)}, \dots, S_i^{(M_i)}\}$, where each sentence represents a specific clinical finding. Here, $S_i^{(k)}$ denotes the $k$-th element in the set of sentences positively paired with image $I_i$. This phrase set was generated using LLM, which was prompted to follow template-based extraction rules, such as \textit{``There [Presence] [Finding] of [Location],''} where square brackets denote template variables. Here, \textit{[Finding]} can include not only a disease term but also detailed visual descriptors, such as ``bilateral mild pleural effusion,'' which provides additional characterization of the disease term ``pleural effusion.''

To better leverage the clinical structure of radiology discourse, we additionally employ LLM to perform abstract-level structuring and sentence mapping. Each extracted phrase is mapped to a structured tuple $(\texttt{Topic}, \texttt{Presence})$, where the `Topic' represents a specific pathology (e.g., `Cardiomegaly') and `Presence' indicates its clinical status (e.g., `Yes', `No', or `Maybe Yes'). From this representation, we generate concise, high-level sentences using the template: \textit{`There [Presence] [Topic].'} These are then integrated as additional positive pairs within the contrastive learning framework. For example, the sentence `There is mild opacity in the bilateral lung base' is simplified to `There is opacity' and utilized as an augmented positive pair. Instead of the original set $T_i$, we use the augmented sentence pair set
$T_i^{\text{aug}} = \{S_i^{(1)}, S_i^{(2)}, \dots, S_i^{(N_i)}\}$, where $N_i \geq M_i$ denotes the number of unique positive sentence pairs associated with $I_i$ after the augmentation process. Finally, the images $\{I_i\}_{i=1}^{B}$ and augmented sentence sets $\{T_i^{\text{aug}}\}_{i=1}^{B}$ are used in the image-sentence contrastive learning.

\vspace{-0.1cm}
\subsection{Feature Extraction}
\label{sec:method_feature_extraction}
\vspace{-0.1cm}
Let $(I_i, T_j^{\text{aug}})$ denote an image–sentence set pair sampled from $i$-th and $j$-th sample of mini-batch $\mathcal{D}_{\text{batch}}$, respectively. 
If the image and sentence set originate from the same sample ($i=j$), the pair is the pair is treated as positive by default; otherwise ($i \neq j$), it is treated as negative.

The image $I_i$ is passed through a visual encoder $f_v$ with a Vision Transformer (ViT)~\citep{dosovitskiy2020vit}-based architecture to obtain a global visual embedding $v_i^{g} \in \mathbb{R}^{D}$ (i.e., the \texttt{[CLS]} token of the final output) and local patch embeddings $v_i^{l} \in \mathbb{R}^{L \times D}$, where $L$ and $D$ denote the number of local patches and the feature dimension, respectively. The encoder $f_v$ consists of a LoRA-finetuned ViT backbone $f_v^{\text{ViT}}$ followed by additional trainable transformer layers $f_v^{\text{train}}$. 

Let $h_i^{(n)} \in \mathbb{R}^{L \times D}$ denote the local patch token sequence produced by the $n$-th layer of the backbone $f_v^{\text{ViT}}$, and let $N$ be its total depth. Since intermediate layers retain complementary information that is partially discarded in the final layer, we do not rely on $h_i^{(N)}$ alone. Instead, we select a set of intermediate layer indices $\mathcal{S}_{\text{layer}} = \{n_1, \dots, n_M\} \subset \{1, \dots, N\}$, transform the representation from each selected layer with a dedicated 2-layer multi-layer perceptron (MLP) followed by LayerNorm~\citep{ba2016layer}, and add their average to the final-layer representation: $$\tilde{h}_i = h_i^{(N)} + \frac{1}{M} \sum_{m=1}^{M} \texttt{LN}_{\text{ViT}}^{(m)}\big(\texttt{MLP}_{\text{ViT}}^{(m)}\big(h_i^{(n_m)}\big) \big),$$ 
where $\texttt{LN}_{\text{ViT}}^{(m)}$ and $\texttt{MLP}_{\text{ViT}}^{(m)}$ denote the LayerNorm and 2-layer MLP associated with the $m$-th selected layer.

The fused token sequence $\tilde{h}_i$ is further refined by the trainable transformer layers $f_v^{\text{train}}$, from whose output the global and local visual representations are read out as $$[v_i^{g}; v_i^{l}] = \texttt{LN}_v\big(f_v^{\text{train}}(\tilde{h}_i)\big), \quad v_i^{l} = \texttt{Concat}\big(\{v_{i,p}\}_{p=1}^{L}\big),$$ where $v_i^{g} \in \mathbb{R}^{D}$ is the output \texttt{[CLS]} token and $v_{i,p} \in \mathbb{R}^{D}$ is the $p$-th patch embedding of image $I_i$. Here, $\texttt{LN}_v$ denotes the LayerNorm applied to the final visual representation, and $\texttt{Concat}(\cdot)$ is the concatenation of the given embeddings along the token dimension.

In parallel, the sentences in $T_j^{\text{aug}}$ are encoded by a bidirectional text encoder $f_t$. Given the $k$-th sentence $S_j^{(k)} \in T_j^{\text{aug}}$ of the $j$-th sample in the batch, its text representation is obtained as $t_j^{(k)} = \texttt{LN}_{t}\big(f_t(S_j^{(k)})\big) \in \mathbb{R}^{D},$
where $\texttt{LN}_{t}$ denotes a LayerNorm applied to the final text representation.

\subsection{Sentence-Conditioned Feature Modulation and Contrastive Loss}
\label{sec:loss_calculation}
Given the patch embeddings and sentence embeddings, we perform image--sentence contrastive learning. Unlike most VL contrastive approaches, which represent an image with a single global embedding, we represent each image by a sentence-conditioned aggregation of its local patch embeddings, in which every patch is weighted by its similarity to the sentence.

Concretely, we first $\ell_2$-normalize the patch embeddings and the sentence embedding, $\bar{v}_{i,p} = v_{i,p} / \|v_{i,p}\|_2$ and $\bar{t}_j^{(k)} = t_j^{(k)} / \|t_j^{(k)}\|_2$, and compute the patch-level similarity $s_{i,j,p}^{(k)} = \langle \bar{v}_{i,p}, \bar{t}_j^{(k)} \rangle$. A softmax over the patches converts these similarities into spatial attention weights:
\begin{equation}
    \label{eq:patch_softmax_calculation}
    a_{i,j,p}^{(k)} =
    \frac{\exp\left(s_{i,j,p}^{(k)} / \tau_a\right)}
    {\sum_{m=1}^{L} \exp\left(s_{i,j,m}^{(k)} / \tau_a\right)},
\end{equation}
where $\tau_a$ is a temperature parameter and $p$ is the patch index. The sentence-conditioned visual feature is the attention-weighted sum of the patch embeddings:
\begin{equation*}
    \tilde{v}_{i,j}^{(k)} = \sum_{p=1}^{L} a_{i,j,p}^{(k)}\, v_{i,p}.
\end{equation*}

In conventional VL contrastive learning, $\tilde{v}_{i,j}^{(k)}$ would be contrasted directly with the sentence embedding. We further refine this visual feature to incorporate sentence semantics through \emph{sentence-conditioned feature modulation}. Specifically, we apply feature-wise linear modulation (FiLM)~\citep{perez2018film} with a residual connection, predicting the scale and shift parameters of LayerNorm as learnable functions of the conditioning feature. This enables sample-specific modulation of the aggregated visual feature, while the residual connection preserves its original visual information.

Specifically, we stack $R=2$ modulation units, each consisting of a feature-wise standardization and a FiLM layer followed by two linear layers. In the $r$-th unit, a 2-layer MLP $\texttt{MLP}_{\text{FiLM}}^{(r)}$ predicts a feature-wise scale $\gamma_{i,j,r}^{(k)}$ and shift $\beta_{i,j,r}^{(k)}$, conditioned on the aggregated patch embedding and the sentence embedding:
\begin{equation*}
    \label{eq:film_params}
    \gamma_{i,j,r}^{(k)}, \beta_{i,j,r}^{(k)} = \texttt{MLP}_{\text{FiLM}}^{(r)}\big(\texttt{Concat}\big(\tilde{v}_{i,j}^{(k)}, t_j^{(k)}\big)\big).
\end{equation*}
Starting from $u_{i,j}^{(k,0)} = \tilde{v}_{i,j}^{(k)}$, the $r$-th unit standardizes its input, modulates it with these sentence-dependent parameters, and then applies two successive linear projections:
\begin{equation*}
    \label{eq:film_unit}
    u_{i,j}^{(k,r)} = W_{r}^{(2)}\Big(W_{r}^{(1)}\big(\gamma_{i,j,r}^{(k)} \odot \texttt{Std}\big(u_{i,j}^{(k,r-1)}\big) + \beta_{i,j,r}^{(k)}\big) + b_{r}^{(1)}\Big) + b_{r}^{(2)}, \quad r = 1, \dots, R,
\end{equation*}
where $\odot$ denotes element-wise multiplication, $\texttt{Std}(x) = (x - \mu(x)) / \sqrt{\sigma^2(x) + \epsilon}$ standardizes a vector $x \in \mathbb{R}^D$ using the mean $\mu(x)$ and variance $\sigma^2(x)$ of its $D$ entries, with a small constant $\epsilon$ for numerical stability (i.e., a LayerNorm without learnable affine parameters, whose role is taken over by $\gamma_{i,j,r}^{(k)}$ and $\beta_{i,j,r}^{(k)}$), and $W_{r}^{(1)}, W_{r}^{(2)} \in \mathbb{R}^{D \times D}$ and $b_{r}^{(1)}, b_{r}^{(2)} \in \mathbb{R}^{D}$ are the parameters of the two linear layers in the $r$-th unit. The output of the last unit is added back to the original aggregated feature through a residual connection, followed by a LayerNorm:
\begin{equation*}
    \label{eq:film_output}
    \hat{v}_{i,j}^{(k)} = \texttt{LN}_{\text{TC}}\big(\tilde{v}_{i,j}^{(k)} + u_{i,j}^{(k,R)}\big),
\end{equation*}
where $\texttt{LN}_{\text{TC}}$ is a LayerNorm applied to the modulated feature. In this way, the stacked module units inject sentence semantics as a learned correction to $\tilde{v}_{i,j}^{(k)}$ rather than replacing it, so the spatially aggregated visual content is retained.

We use $\hat{v}_{i,j}^{(k)}$ as the visual side of the image--sentence contrastive loss and $\ell_2$-normalize it to obtain $\bar{v}_{i,j}^{(k)}$. The similarity logit is then the temperature-scaled cosine similarity between the sentence-conditioned visual feature and the sentence embedding:
\begin{equation}
    \label{eq:logit_calculation}
    z_{i,j}^{(k)} = \langle \bar{v}_{i,j}^{(k)}, \bar{t}_{j}^{(k)} \rangle / \tau_l,
\end{equation}
where $\tau_l$ is the logit temperature and $\bar{t}_{j}^{(k)}$ is the $\ell_2$-normalized text embedding.

Using these logits, we compute a contrastive loss between the images and sentences in each batch. Since a single image can have multiple positive sentences, we adopt the multi-positive noise-contrastive estimation (MP-NCE) loss~\citep{lee2022uniclip} with softmax-based competition:
\begin{equation*}
\label{eq:infonce}
\mathcal{L}_{I}\!=\!-\frac{1}{N_T}\! \sum_{i=1}^{B}\!\sum_{n=1}^{N_i} \log \frac{\exp(z_{i,i}^{(n)})} {\exp(z_{i,i}^{(n)})\!+\!\sum_{j \neq i}^{B}\sum_{m=1}^{N_j}\exp(z_{i,j}^{(m)})},
\end{equation*}
where $N_j$ denotes the number of positive pair sentences for the $j$-th sample and $N_T$ is the total number of sentences in the batch. That is, each positive sentence competes only against the negative sentences from other images in the batch, and the per-pair losses are averaged over all positive pairs.

For each finding sentence, we apply a text-to-image InfoNCE loss~\citep{oord2018representation}, defined as
\begin{equation*}
\mathcal{L}_{T}
=
-\frac{1}{N_T}
\sum_{i=1}^{B}\sum_{n=1}^{N_i}
\log
\frac{\exp(z_{i,i}^{(n)})}
{\exp(z_{i,i}^{(n)})+\sum_{j\neq i}^{B}\exp(z_{j,i}^{(n)})}.
\end{equation*}
Each sentence is paired with exactly one image, and the standard InfoNCE loss applies directly in this direction. The contrastive loss $\mathcal{L}_{\mathrm{con}}$ is defined as the average of the image-to-text and text-to-image contrastive losses, denoted as $\mathcal{L}_{\mathrm{con}}=(\mathcal{L}_{I}+\mathcal{L}_{T}) / 2$.

\subsection{False Negative Loss}
\label{sec:false_neg_mitigation}
However, treating all unpaired examples within a batch as strict negatives
unfairly penalizes \textit{false negatives}, i.e., image--sentence pairs drawn
from different studies that nevertheless describe the same pathology. To
mitigate this, we introduce a false-negative identification strategy.
Specifically, if the $k$-th sentence of the $j$-th sample, $S_j^{(k)}$, also
appears in the augmented text set $T_i^{\mathrm{aug}}$ of the $i$-th sample, we
treat the pair $(I_i, S_j^{(k)})$ as a false negative. Accordingly, given a
mini-batch of size $B$, we define the set of false-negative pairs $\mathcal{F}$ as
\[
\mathcal{F} = \bigl\{\, (i, j, k) \;|\;
i \neq j \text{ and } S_j^{(k)} \in T_i^{\mathrm{aug}};\ k=1,2,\dots,N_j \bigr\}_{i,j=1}^{B}.
\]

Rather than pulling the global image embedding toward every false-negative
sentence, we align only the image regions most relevant to that sentence. For each
false-negative pair $(i,j,k) \in \mathcal{F}$, we compute the cosine similarity
$s_{i,j,p}^{(k)} = \cos\!\bigl(v_{i,p}, t_j^{(k)}\bigr)$
between every patch and sentence, and select the top-$M$\% patch indices
$\mathcal{P}_{i,j}^{(k)}$. The false-negative loss $\mathcal{L}_{fn}$ then aims to maximize the cosine similarity between the selected patches and the sentence, averaged over all
false-negative pairs:
\begin{equation*}
\mathcal{L}_{fn} =
\frac{1}{|\mathcal{F}|}
\sum_{(i,j,k) \in \mathcal{F}}
\frac{1}{|\mathcal{P}_{i,j}^{(k)}|}
\sum_{p \in \mathcal{P}_{i,j}^{(k)}}
\bigl(1 - s_{i,j,p}^{(k)}\bigr).
\end{equation*}
This auxiliary loss attracts only the top-$M\%$ highest-similarity patches toward each false-negative sentence identified through cross-study matching, counteracting repulsion through selective local alignment. It directly supervises the selected patches while preserving the global contrastive objective and its pair assignments. Thus, our false-negative mitigation does not remove false negatives from the contrastive loss; instead, it adds patch-level supervision that aligns each image with the clinical statements it shares with other studies.

The total training loss $\mathcal{L}_{train}$ is
then defined as
$$\mathcal{L}_{train} = \lambda_{con}\mathcal{L}_{con} + \lambda_{fn}\mathcal{L}_{fn},$$
where $\lambda_{con}$ and $\lambda_{fn}$ are the coefficients for each loss term.

\subsection{Multi-Task Zero-Shot Inference}
\label{sec:method_inference}

After pretraining, SentZero supports zero-shot generalization across diverse downstream tasks by leveraging the learned alignment between visual features and text prompts (bottom of Fig.~\ref{fig:main}).

\vspace{-0.2cm}
\paragraph{Zero-Shot Classification.} To perform classification for a specific pathology, we construct a text prompt $S_{test}$ using the clinical template (e.g., \textit{``There is [Finding]''}). For a given test image $I_{test}$, we first compute the sentence-specific attended visual feature $\bar{v}_{test}$ and the text representation $\bar{t}_{test}$ as described in Sec.~\ref{sec:loss_calculation}. The similarity logit is computed as $z = \langle \bar{v}_{test}, \bar{t}_{test} \rangle / \tau_l$ as in~\eqref{eq:logit_calculation}, and converted into a probability via a sigmoid function, denoted as $\hat{p} = \sigma(z).$ This probability $\hat{p}$ represents the model's confidence in the presence of the finding described by the prompt. 

\vspace{-0.3cm}
\paragraph{Zero-Shot Grounding.} To localize findings, we utilize the spatial attention weights derived from~\eqref{eq:patch_softmax_calculation}. For each local patch $p \in \{1, \dots, L\}$, the attention weight $a_{p}$ is calculated as:
\begin{equation*}
    a_{p} = \frac{\exp(s_{p} / \tau_a)}{\sum_{m=1}^L \exp(s_{m} / \tau_a)},
\end{equation*}
where $s_{p} = \langle \bar{v}_{p}, \bar{t}_{test} \rangle$ is the patch-level similarity score. The resulting set of weights $\{a_{p}\}_{p=1}^{L}$ forms an attention map $A \in \mathbb{R}^{\sqrt{L} \times \sqrt{L}}$. 

This map is then reshaped and bilinearly upsampled to the original image resolution, yielding the final Softmax Attention Map $\hat{A} = \texttt{bilinear}(A)$, where $\texttt{bilinear}$ denotes bilinear interpolation. The acquired attention map $\hat{A}$ provides a relative distribution of importance across the image, where the values sum to one across the spatial domain. For grounding, the peak of this distribution (the maximum value in $\hat{A}$) is used to identify the most likely location of the finding.

%% file: sections/5_experiments.tex
\section{Experiments}
\label{sec:experimental_setup}
\subsection{Experimental Settings}
\vspace{-0.2cm}
\paragraph{Datasets.}
We use MIMIC-CXR~\citep{mimic_cxr_physionet,johnson2019mimic} for vision--language pretraining, including all frontal and lateral views and following the official dataset split. MIMIC-CXR consists of image--report pairs from 377K chest X-ray images, 227K radiographic studies, and 65,379 patients. For report text, we use the \textit{findings} section for phrase extraction and discard studies without extracted finding sentences. To ensure a fair comparison with RadZero~\citep{park2025radzero}, we adopt its phrases extracted by \texttt{LLaMA3-70B-Instruct} and further apply abstract-level sentence mapping using \texttt{Qwen3-Next-80B}. The detailed instruction prompt is provided in the Appendix~\ref{appendix:dataset}.

\vspace{-0.1cm}
For evaluation, we follow the multi-task zero-shot CXR benchmarks that have been commonly used in prior works~\citep{huang2021gloria, zhang2023knowledge, lai2024carzero, wu2023medklip, park2025radzero}. Zero-shot classification is evaluated on Open-I~\citep{demner2016preparing}, ChestXray14~\citep{wang2017chestx}, PadChest~\citep{bustos2020padchest}, ChestXDet10~\citep{liu2020chestx}, and CheXpert~\citep{irvin2019chexpert}; and grounding is evaluated on ChestXDet10 and MS-CXR~\citep{boecking2022making}. All classification datasets are used for multi-label disease classification, while the grounding datasets provide text--bounding box pairs for diverse disease expressions. Additional dataset details, including the number of samples, are provided in the Appendix~\ref{appendix:dataset}.

\vspace{-0.3cm}
\paragraph{Baseline Models and Evaluation Metrics.}
Although many VL pretraining methods have been proposed in the CXR domain, only a few support zero-shot transfer. We therefore compare our model with existing VL pretraining methods that can be directly applied to zero-shot downstream tasks: GLoRIA~\citep{huang2021gloria}, BioViL-T~\citep{bannur2023learning}, MedKLIP~\citep{wu2023medklip}, KAD~\citep{zhang2023knowledge}, CARZero~\citep{lai2024carzero}, RadZero~\citep{park2025radzero}, CoNNs~\citep{lian2026concept}, and GLINT~\citep{park2026glint}. For GLINT, we use the ViT-B variant with a 512$\times$512 input resolution, which is comparable to our model in both backbone size and input size.

\vspace{-0.1cm}
We follow the evaluation protocols used in these prior studies~\citep{huang2021gloria, zhang2023knowledge, lai2024carzero, wu2023medklip, park2025radzero}. For zero-shot classification, we report the area under the receiver operating characteristic curve (\textbf{AUROC}) on multi-label test datasets. In zero-shot grounding, we use \textbf{pointing game accuracy}~\citep{zhang2018top, lai2024carzero, park2025radzero}, which measures whether the spatial location with the highest model response falls within the corresponding ground-truth bounding box.

\vspace{-0.3cm}
\paragraph{Implementation Details.}
The vision encoder consists of a pretrained RAD-DINO model~\citep{perez2025exploring} with a LoRA adapter (rank=16, and alpha=48) followed by four trainable transformer blocks. The input image size is $518 \times 518$, and the resulting feature map has a spatial resolution of $37 \times 37$, corresponding to 1369 patches. For the text encoder, we use MPNet (\texttt{all-mpnet-base-v2})~\citep{reimers2019sentence, song2020mpnet}. For intermediate-layer feature selection in the vision encoder, we extract features from the 3rd, 6th, and 9th layers of the ViT backbone and fuse them with the final-layer hidden representations. 

\vspace{-0.1cm}
For loss computation, all temperature hyperparameters are set to $\tau_a = \tau_l = 0.1$. The loss coefficients are set to $\lambda_{con} = 1$ and $\lambda_{fn} = 0.01$, with a patch sampling ratio of $M=20\%$ for false negative pairs. The model is trained for up to 20 epochs using the AdamW optimizer with an initial learning rate of $1 \times 10^{-4}$ and a batch size of 256. Training is performed on four NVIDIA H200 GPUs with DeepSpeed ZeRO Stage 2 for memory efficiency, using a \texttt{WarmupCosineLR} scheduler. We apply early stopping with a patience of 3 epochs and select the checkpoint with the lowest validation loss as the final model; training completes in approximately 5 hours. All ablation studies are conducted under the same fixed random seed. Additional results of hyperparameter tuning are described in Appendix~\ref{appendix:additional experiment}.

\vspace{-0.2cm}
\subsection{Main Results}
\vspace{-0.1cm}
Table~\ref{tab:multi_task_results} presents the zero-shot performance of our method and prior baselines across a range of downstream vision tasks. Our method achieves the best results across multiple multi-label datasets and task types. While CARZero~\citep{lai2024carzero} attains the highest AUROC on CheXpert, this dataset is relatively small (500 cases), and CARZero does not maintain comparable performance on the larger classification datasets, which contain thousands to tens of thousands of samples (see Appendix~\ref{appendix:dataset} for detailed dataset profiles). Our proposed model also demonstrates improvements on localization tasks. On ChestXDet10 and MS-CXR, it outperforms the strongest baseline by 7.6 and 3.6 percentage points in pointing game accuracy, respectively.

\vspace{-0.1cm}
\input{table/main_table}
\vspace{-0.1cm}

\subsection{Ablation Studies on Model Components}
\vspace{-0.2cm}
Table~\ref{tab:ablation_main} reports the ablation results for each component of SentZero. Starting from the baseline without any proposed component (first row), multi-layer feature aggregation improves performance, most notably on the grounding tasks (second row). Adding each of the remaining components individually on top of it (third to fifth rows) further improves classification performance, and combining all components yields the best or near-best results on most datasets. Although the text-conditioned residual connection alone achieves higher scores on PadChest and CheXpert, the full model outperforms it on the other five benchmarks by larger margins.

To assess the effect of text conditioning, we remove the text embeddings from the feature modulation units, replacing them with duplicated aggregated patch features to keep the parameter count and projection dimension unchanged. Incorporating text features improves performance on most datasets (See Table~\ref{tab:text_feature_ablation} in Appendix~\ref{appendix:additional experiment}).

\input{table/ablation_main}

\vspace{-0.1cm}
\subsection{Comparison on False Negative Handling Strategy}
\label{sec:result_false_neg_comparison}
\vspace{-0.2cm}
We compare our strategy with two simple alternatives that directly modify the pair assignments in the contrastive loss $\mathcal{L}_{con}$: (1) \emph{false-negative masking}, which removes false-negative pairs from the denominator, and (2) \emph{false-negative transition}, which relabels them as positive pairs. Prior work such as CoNNs~\citep{lian2026concept} employs both operations, selectively applying them according to the presence status of each finding. Table~\ref{tab:fn_comparison} presents the results, where `None' denotes training without any false-negative handling (the sixth row of Table~\ref{tab:ablation_main}). Both alternatives underperform our strategy on most benchmarks and, in most cases, also underperform the `None' baseline. In particular, relabeling false negatives as positives causes large performance drops across multiple datasets, suggesting that directly modifying contrastive pair assignments can impair learning in this setting. These findings support our auxiliary loss, which selectively attracts the highest-similarity patches toward each false-negative sentence while preserving the original contrastive pair assignments.

\input{table/false_neg_comparison}

\vspace{-0.1cm}
\subsection{Visualization of Attention Map}
\begin{figure}[h!]
\centering
\vspace{-0.4cm}
\begin{subfigure}{0.45\textwidth}
    \centering
    \includegraphics[width=\linewidth]{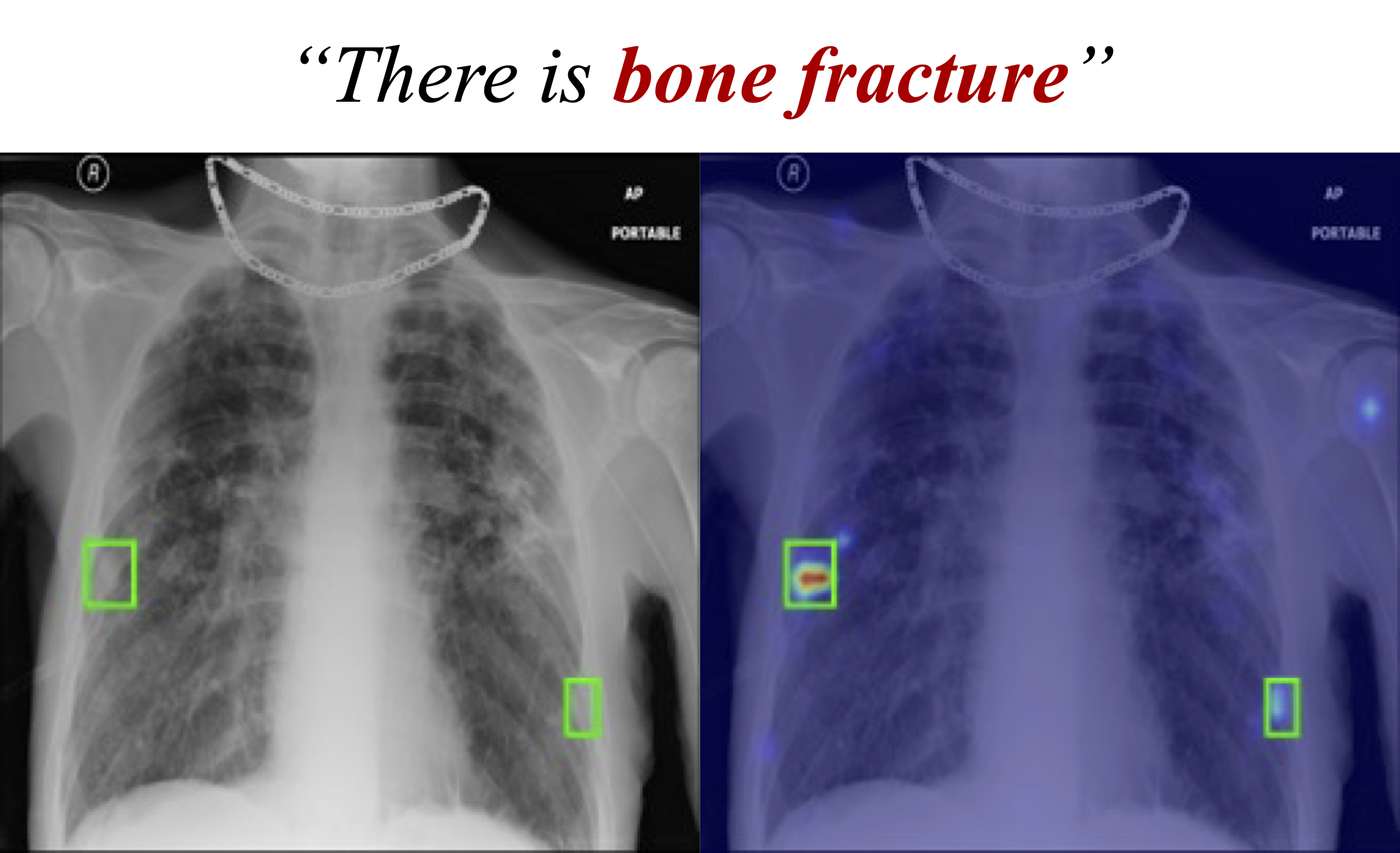}
\end{subfigure}
\hspace{0.2cm}
\begin{subfigure}{0.45\textwidth}
    \centering
    \includegraphics[width=\linewidth]{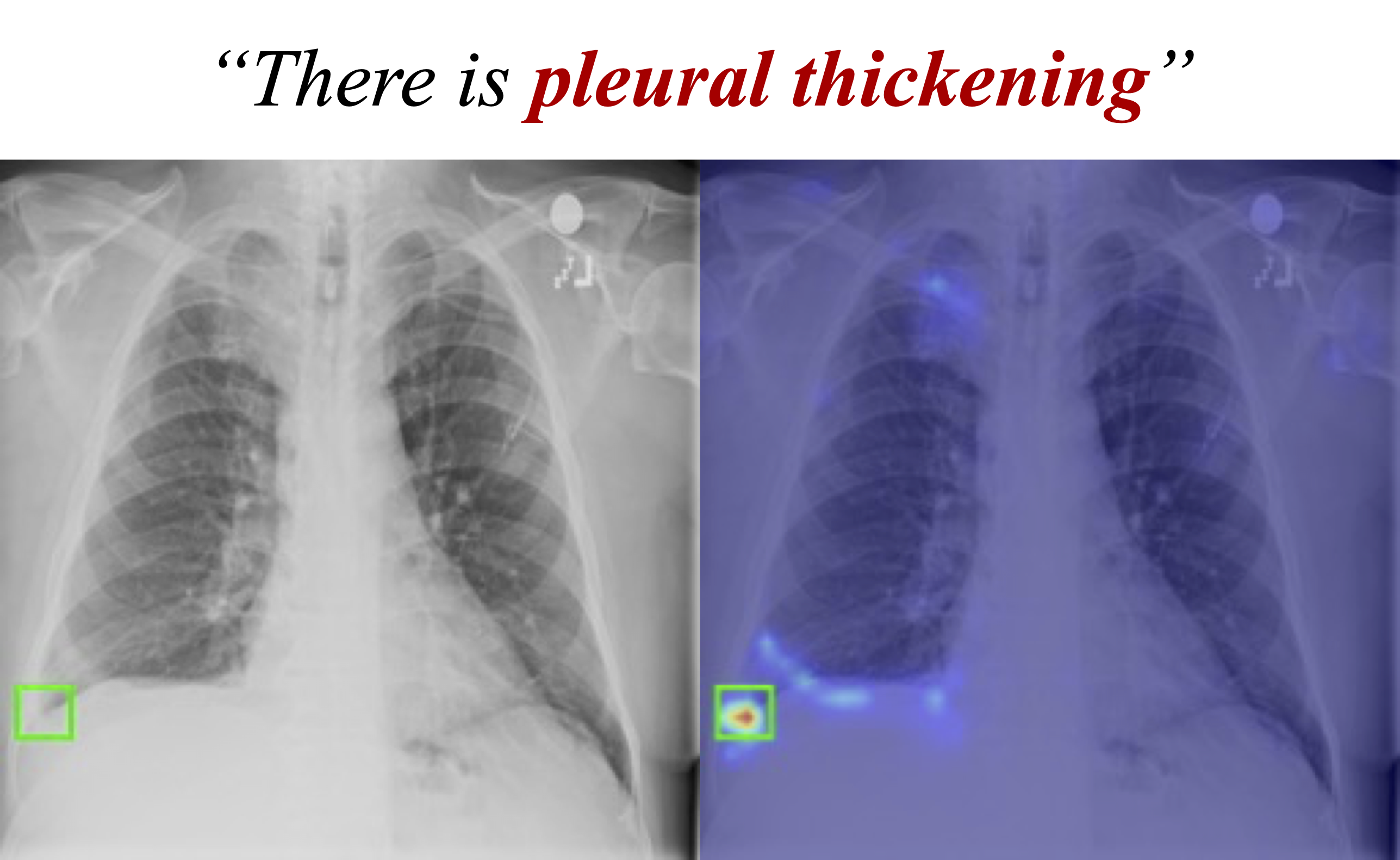}
\end{subfigure}
\vspace{-0.2cm}
\caption{Examples of attention heatmaps for relatively small lesions from the ChestXDet10 dataset. Each example shows the original image and its attention map for the given text prompt. Green boxes are the ground-truth bounding boxes.}
\label{fig:attn_map_examples}
\vspace{-0.3cm}
\end{figure}

\vspace{-0.1cm}
Fig.~\ref{fig:attn_map_examples} visualizes attention maps for examples from the ChestXDet10 dataset, which provides bounding box annotations grounded to text phrases. As shown in the figure, the attention maps accurately highlight relevant pathological regions, including small focal lesions occupying only a limited area. These visualization results suggest that the attention maps produced by SentZero have strong potential for zero-shot visual grounding. Additional examples of visualized attention maps are in Appendix~\ref{appendix:visualization}.

%% file: table/main_table.tex
\begin{table*}[h]
\centering
\small
\vspace{-0.1cm}
\caption{Zero-shot performance comparison results. The first header row indicates the task type and corresponding evaluation metric. The best and second-best results are shown in \textbf{bold} and \underline{underlined}, respectively. CXR14 and CXD10 refer to the ChestXray14 and ChestXDet10 datasets, respectively.}
\vspace{-0.2cm}
\resizebox{\textwidth}{!}{%
\begin{tabular}{l|c|ccccc|cc}
\toprule
\multirow{3}{*}{Method} & \multirow{3}{*}{Venue} & \multicolumn{5}{c|}{Classification} & \multicolumn{2}{c}{Grounding} \\
 &  & \multicolumn{5}{c|}{(AUROC)} & \multicolumn{2}{c}{(Pointing Acc.)} \\
\cmidrule(lr){3-7} \cmidrule(lr){8-9}
 &  & OpenI & CXR14 & PadChest & CXD10 & CheXPert & CXD10 & MS-CXR \\ \hline 
GLoRIA & ICCV'21 & 0.589 & 0.610 & 0.565 & 0.645 & 0.750 & 0.367 & - \\
BioViL-T & CVPR'23 & 0.702 & 0.729 & 0.655 & 0.708 & 0.789 & 0.351 & 0.719 \\
MedKLIP & ICCV'23 & 0.759 & 0.726 & 0.629 & 0.713 & 0.879 & 0.481 & 0.407 \\
KAD & Nat. Comm.'23 & 0.807 & 0.789 & 0.750 & 0.735 & 0.905 & 0.391 & - \\
CARZero & CVPR'24 & 0.838 & 0.811 & 0.810 & 0.796 & \textbf{0.923} & 0.543 & 0.749 \\
RadZero & NeurIPS'25 & 0.847 & 0.804 & 0.841 & 0.787 & 0.900 & 0.622 & 0.844 \\
CoNNs & MICCAI'26 & \underline{0.871} & \underline{0.819} & 0.835 & \underline{0.825} & \underline{0.920} & \underline{0.656} & 0.872 \\
GLINT & NeurIPS'26 & \underline{0.871} & 0.817 & \underline{0.853} & 0.812 & 0.918 & 0.625 & \underline{0.886} \\
\midrule 
% SentZero (224px) & - & 0.865 & 0.828 & 0.856 & 0.810 & 0.892 & 0.579 & 0.880 \\
SentZero (Ours) & - & \textbf{0.889} & \textbf{0.839} & \textbf{0.861} & \textbf{0.843} & 0.904 & \textbf{0.732} & \textbf{0.922} \\
\bottomrule
\end{tabular}%
}
\label{tab:multi_task_results}
\vspace{-0.3cm}
\end{table*}

%% file: table/ablation_main.tex
\begin{table*}[t]
\centering
\small
\caption{Ablation results for the model components. MF -- Multi-Layer Feature Aggregation; ALM -- Abstract-Level Mapping; TC -- Text Conditioned Residual Connection; FN -- False Negative Mitigation. Best results in each column are in \textbf{bold}.}
\vspace{-0.2cm}
\resizebox{0.93\textwidth}{!}{%
\begin{tabular}{cccc|ccccc|cc}
\toprule
\multicolumn{4}{c|}{Component}
& \multicolumn{5}{c|}{Classification}
& \multicolumn{2}{c}{Grounding} \\
\cmidrule(lr){1-4}
\cmidrule(lr){5-9}
\cmidrule(lr){10-11}
MF & ALM & TC & FN
& OpenI & CXR14 & PadChest & CXD10 & CheXPert
& CXD10 & MS-CXR \\
\midrule
& & & & 0.8747 & 0.8140 & 0.8592 & 0.8090 & 0.9080 & 0.6513 & 0.8922 \\
\checkmark & & & & 0.8741 & 0.8164 & 0.8559 & 0.8093 & 0.9109 & 0.6655 & \textbf{0.9222} \\
\checkmark & \checkmark & & & 0.8856 & 0.8323 & 0.8635 & 0.8395 & 0.8978 & 0.6811 & 0.9162 \\
\checkmark & & \checkmark & & 0.8828 & 0.8236 & \textbf{0.8652} & 0.8226 & \textbf{0.9162} & 0.7032 & 0.8922 \\
\checkmark & & & \checkmark & 0.8774 & 0.8239 & 0.8607 & 0.8230 & 0.9155 & 0.6556 & 0.8862 \\
\checkmark & \checkmark & \checkmark & & 0.8858 & 0.8376 & 0.8627 & 0.8431 & 0.9047 & 0.7121 & 0.9042 \\
\checkmark & \checkmark & \checkmark & \checkmark & \textbf{0.8891} & \textbf{0.8389} & 0.8609 & \textbf{0.8433} & 0.9037 & \textbf{0.7315} & \textbf{0.9222} \\
\bottomrule
\end{tabular}%
}
\label{tab:ablation_main}
\vspace{-0.2cm}
\end{table*}

%% file: table/false_neg_comparison.tex
\begin{table*}[t]
\centering
\centercaption
\small
\caption{Comparison of false negative mitigation strategies. Best mitigation results in each column are in \textbf{bold}.}
\vspace{-0.2cm}
\resizebox{0.93\textwidth}{!}{%
\begin{tabular}{l|ccccc|cc}
\toprule
\multirow{2}{*}{Method}
& \multicolumn{5}{c|}{Classification}
& \multicolumn{2}{c}{Grounding} \\
\cmidrule(lr){2-6}
\cmidrule(lr){7-8}
& OpenI & CXR14 & PadChest & CXD10 & CheXPert
& CXD10 & MS-CXR \\
\midrule
None & 0.8858 & 0.8376 & 0.8627 & 0.8431 & 0.9047 & 0.7121 & 0.9042 \\
\midrule
False Negative Masking    & 0.8815 & 0.8334 & \textbf{0.8642} & \textbf{0.8444} & 0.8981 & 0.6923 & 0.8982 \\
False Negative Transition & 0.8492 & 0.8139 & 0.7904 & 0.7951 & \textbf{0.9067} & 0.5138 & 0.7066 \\
Ours                      & \textbf{0.8891} & \textbf{0.8389} & 0.8609 & 0.8433 & 0.9037 & \textbf{0.7315} & \textbf{0.9222} \\
\bottomrule
\end{tabular}%
}
\label{tab:fn_comparison}
\vspace{-0.5cm}
\end{table*}

%% file: sections/6_conclusion.tex
\section{Conclusion}
\label{sec:conclusion}
In this work, we presented SentZero, a sentence-centric vision–language pretraining framework built for the semantic and structural complexity of CXR reports. SentZero advances sentence-level pretraining through abstract-level mapping, selective patch-level attraction for shared clinical statements, and sentence-conditioned residual modulation of visual features, complementing prior work on structured supervision and contrastive pair relabeling. SentZero outperforms prior multi-task CXR models across diverse zero-shot tasks, showing that explicitly modeling semantic hierarchy and cross-report expression overlap is key to robust, generalizable image–sentence alignment. More broadly, SentZero lays a scalable foundation for semantically aware medical vision–language pretraining, connecting raw clinical text to fine-grained visual understanding.

%% file: sections/7_appendix.tex
\section{Additional Dataset Details}
\label{appendix:dataset}
\subsection{Dataset Profiles}
We evaluate our model on a diverse set of public CXR benchmarks spanning classification, localization, and phrase grounding tasks. We verified that these evaluation images were excluded from both SentZero training and the pretraining data of the RAD-DINO checkpoint used in our experiments.

\vspace{-0.5cm}
\paragraph{Classification Datasets.}
OpenI contains 7,470 CXR images paired with 3,851 radiology reports and multi-label annotations for 18 disease categories. ChestXray14 provides an official test set of 22,433 images annotated with 14 disease labels. PadChest comprises 160,868 CXR images from 67,000 patients and provides 192 labels with a highly long-tailed distribution. Following prior studies~\citep{lai2024carzero, park2025radzero}, we use the subset of 39,053 samples annotated by board-certified radiologists. CheXpert includes a test set of images from 500 patients, labeled by five board-certified radiologists. Following~\citep{lai2024carzero}, we evaluate classification performance on five observations: atelectasis, cardiomegaly, consolidation, edema, and pleural effusion.

\vspace{-0.5cm}
\paragraph{Grounding Datasets.}
For visual grounding evaluation, we use ChestXDet10 and MS-CXR. ChestXDet10 is a subset of ChestXray14 and provides 542 official test images with bounding box annotations for 10 disease categories. Since ChestXDet10 also includes disease-level labels, we additionally evaluate classification performance on this dataset. MS-CXR contains 1,153 image--phrase--bounding box triplets derived from MIMIC-CXR. Because each bounding box is linked to a specific phrase from the corresponding radiology report, MS-CXR enables fine-grained phrase grounding evaluation. For a fair comparison, we follow~\citep{chen2023medical} and evaluate on the released test set of 167 images.

\subsection{Instructions for LLM-Based Abstract-Level Mapping}
\begin{figure*}[h]
    \centering
    \centercaption
    \includegraphics[width=\textwidth]{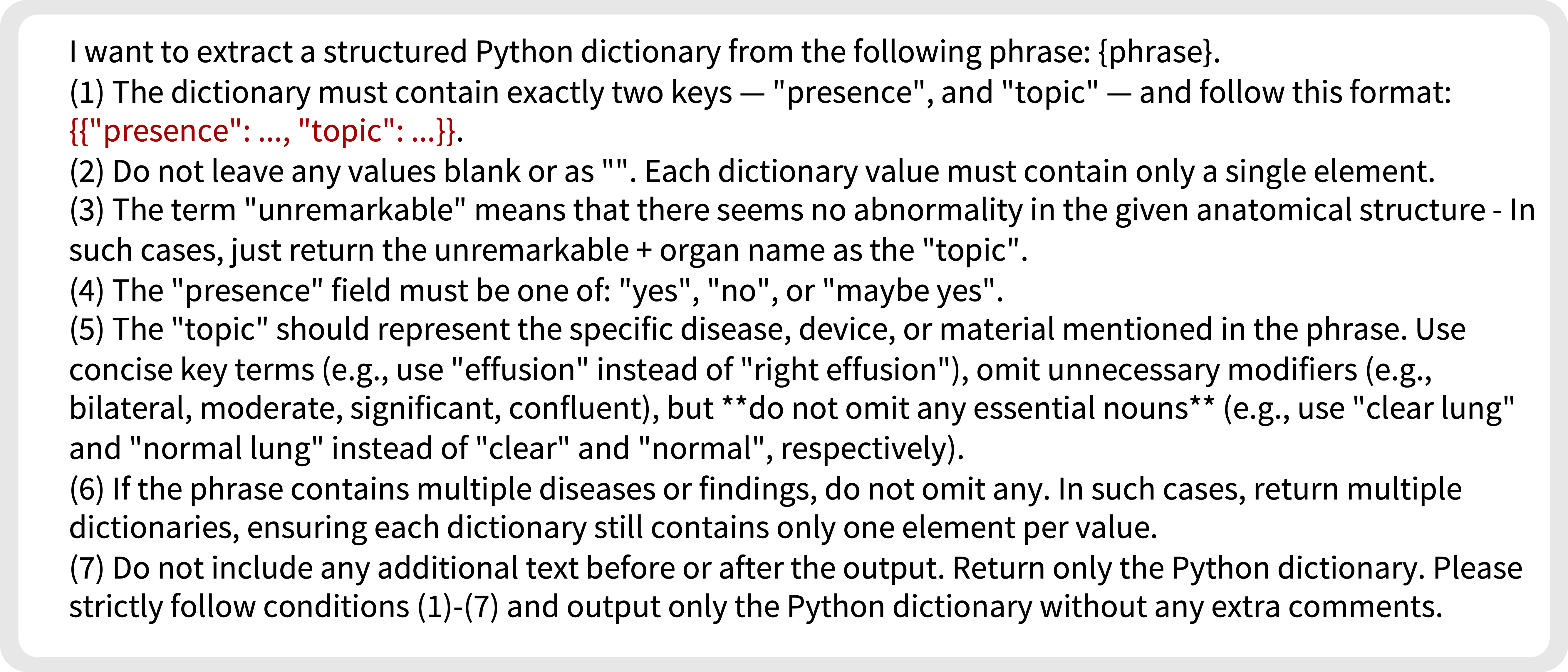}
    \vspace{-0.3cm}
    \caption{Instruction prompt used to extract structured tuples for abstract-level sentence mapping.}
    \label{fig:prompt_example}
    \vspace{-0.6cm}
\end{figure*}

As described in Sec.~\ref{sec:method_phrase_extraction}, we instruct \texttt{Qwen3-Next-80B} to additionally extract a structured tuple containing topic and presence information for abstract-level sentence mapping. The instruction used for this process is shown in Fig.~\ref{fig:prompt_example}. We filter the generated outputs based on their format, and revise the few samples that do not conform to the Python dictionary format.

\newpage
\section{Additional Experiments}
\label{appendix:additional experiment}
Table~\ref{tab:text_feature_ablation} shows the ablation results of text feature conditioning in the feature modulation unit. Text conditioning improves performance on most datasets, indicating the benefit of incorporating textual context into feature modulation.

\input{table/ablation_text_feature}
Tables~\ref{tab:hyperparam_fn_coeff} and~\ref{tab:hyperparam_fn_ratio} report the effects of varying the false-negative loss coefficient and patch sampling ratio, respectively. In Table~\ref{tab:hyperparam_fn_ratio}, the coefficient is fixed at $\lambda_{fn}=0.01$. For the results reported in the main table, we use $\lambda_{fn}=0.01$ and a patch sampling ratio of 20\%.

\input{table/hyperparam_false_neg_w}
\input{table/hyperparam_false_neg_k}

\newpage
\section{Additional Visualization Examples}
\label{appendix:visualization}
\begin{figure*}[h]
    \centering
    \centercaption
    \begin{minipage}{0.4\linewidth}
        \centering
        \includegraphics[width=\linewidth]{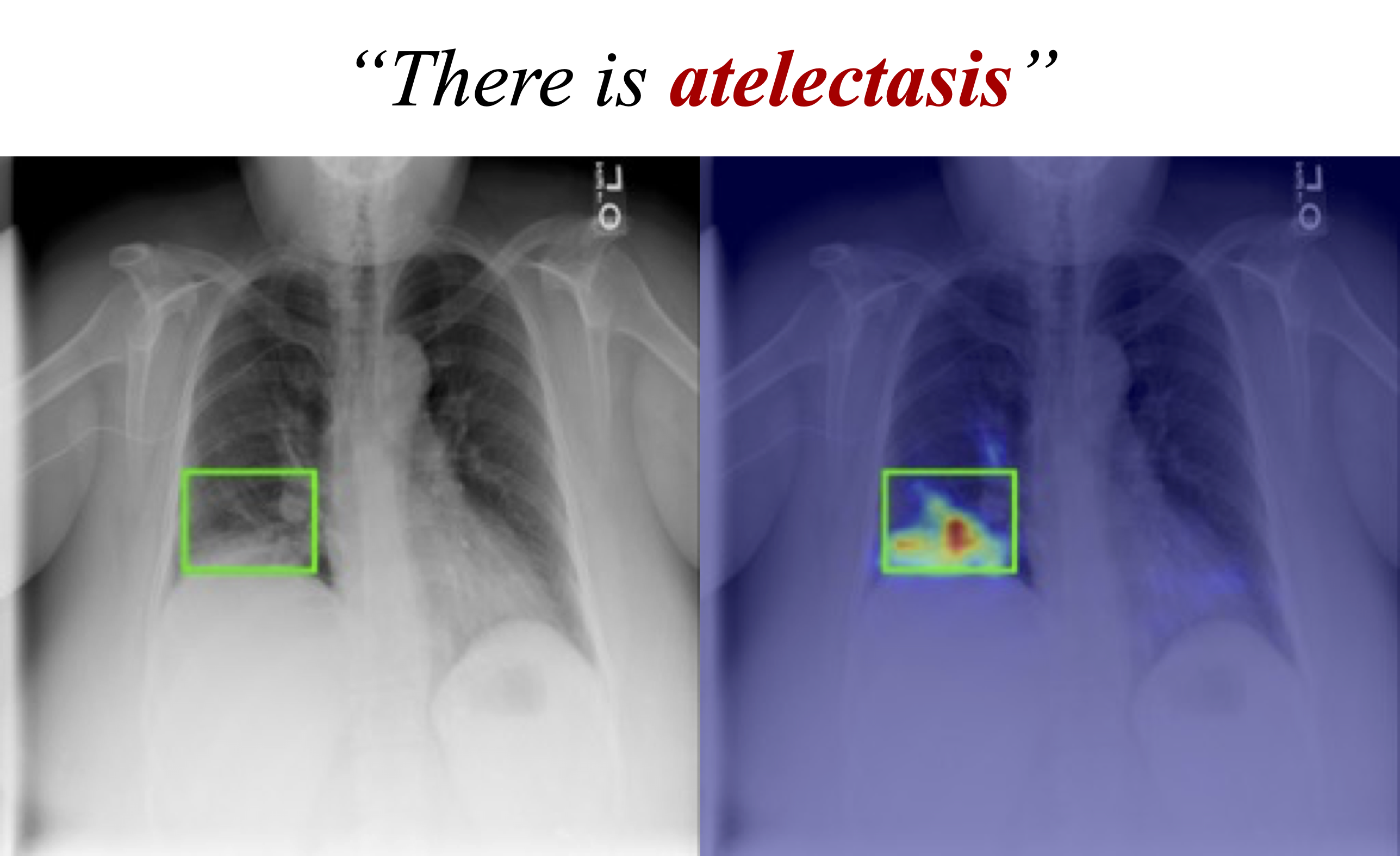}
    \end{minipage}
    \hspace{0.02\linewidth}% adjust as desired
    \begin{minipage}{0.4\linewidth}
        \centering
        \includegraphics[width=\linewidth]{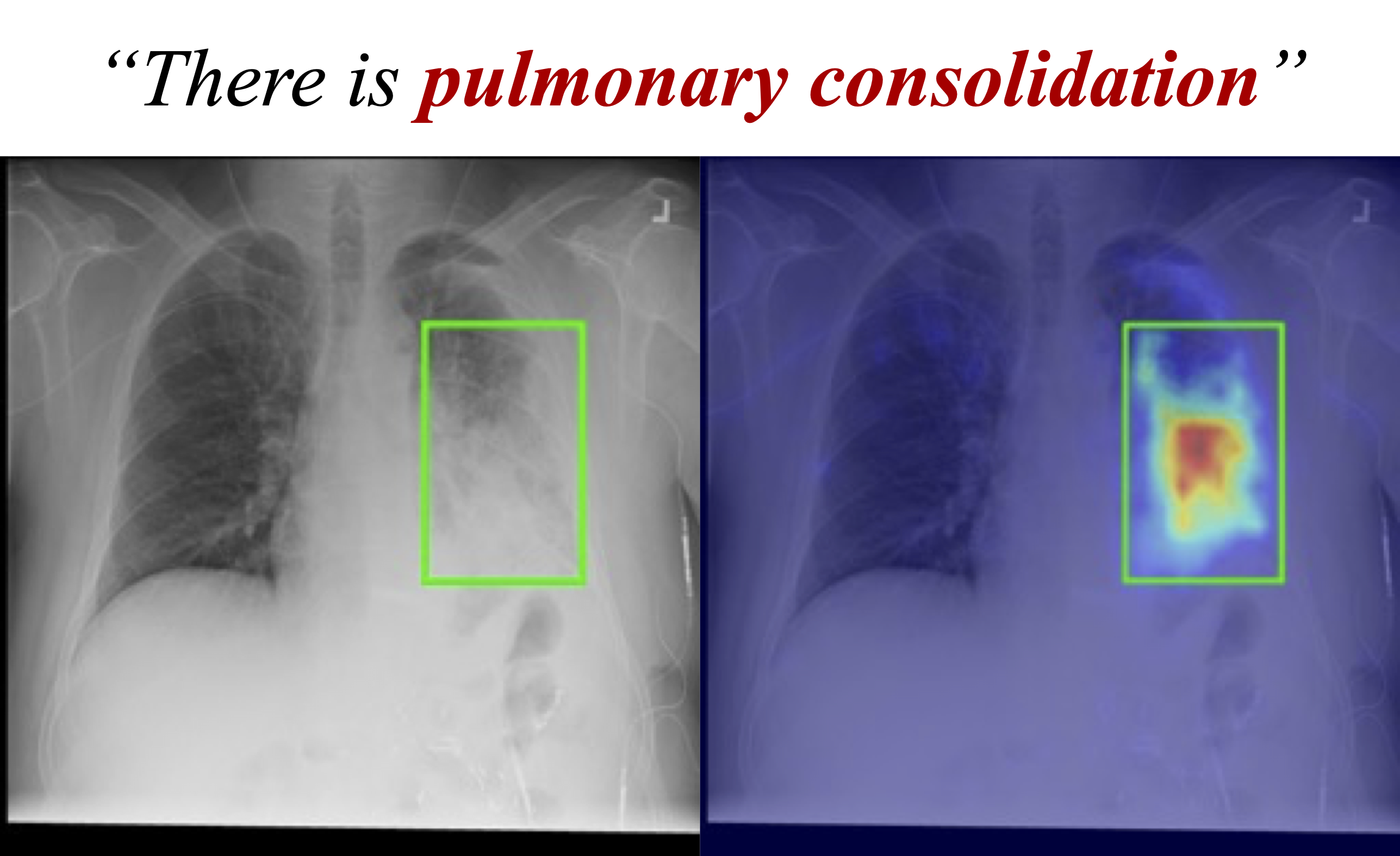}
    \end{minipage}

    \vspace{0.1cm}

    \begin{minipage}{0.4\linewidth}
        \centering
        \includegraphics[width=\linewidth]{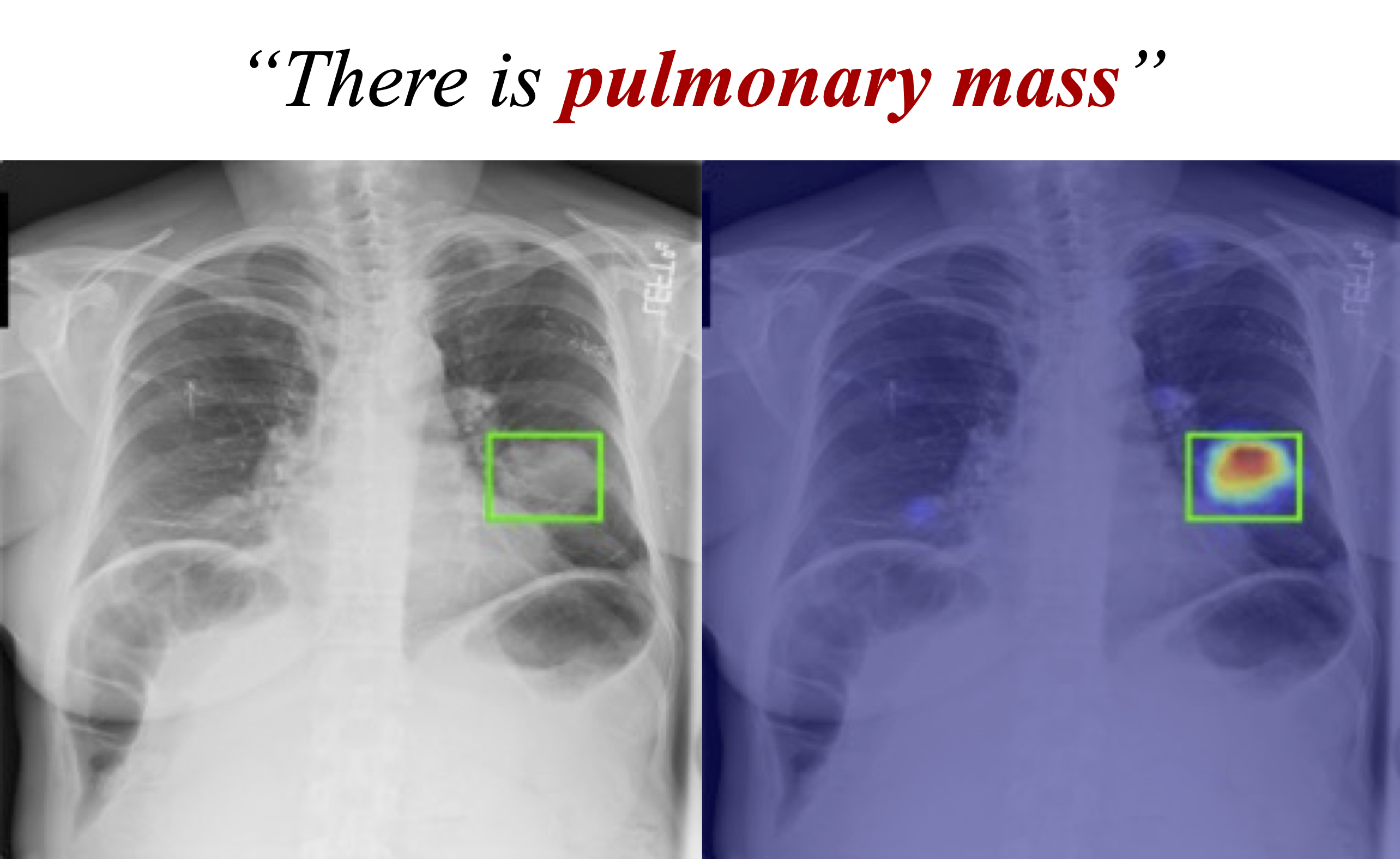}
    \end{minipage}
    \hspace{0.02\linewidth}% adjust as desired
    \begin{minipage}{0.4\linewidth}
        \centering
        \includegraphics[width=\linewidth]{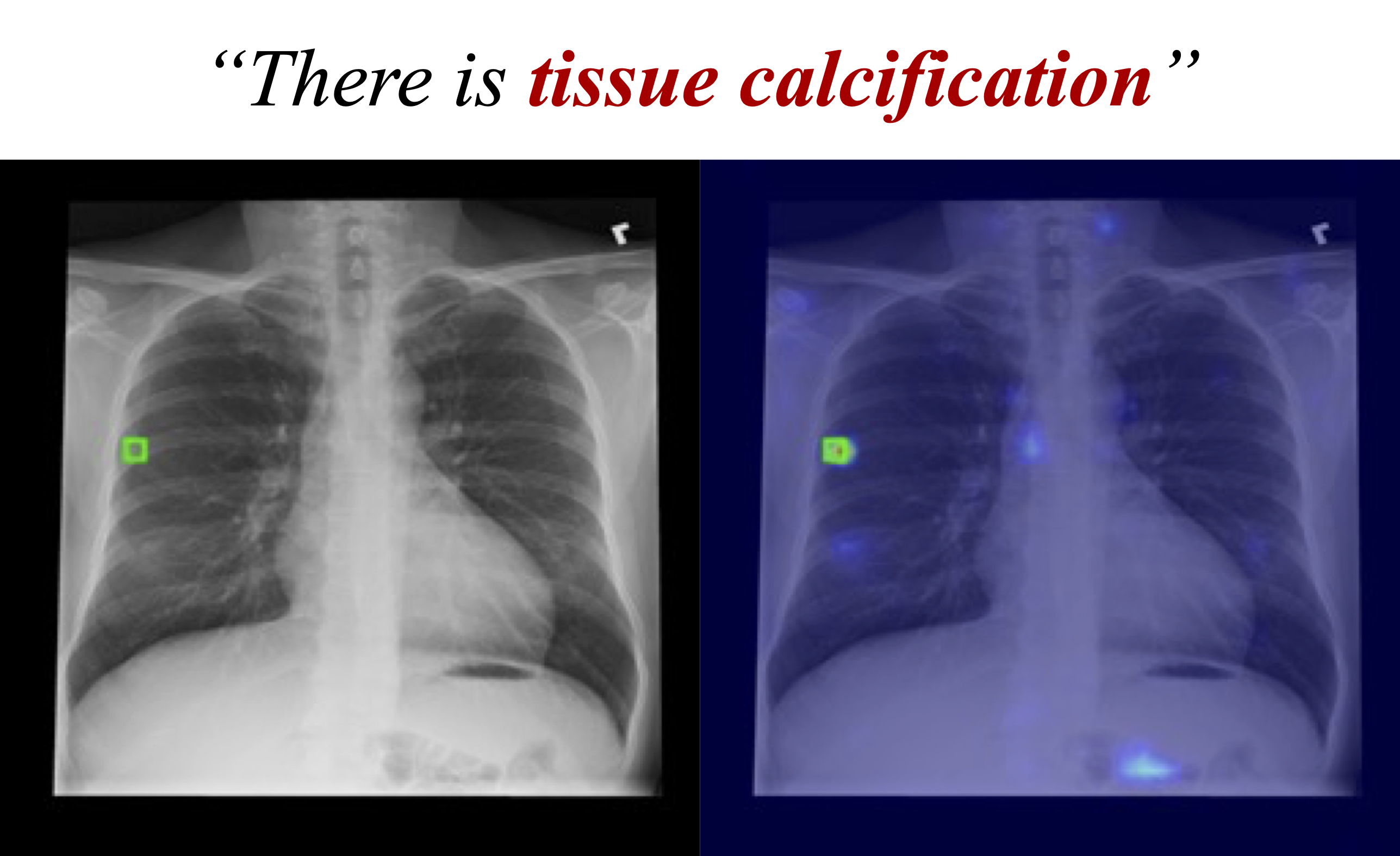}
    \end{minipage}
    \vspace{-0.2cm}
    \caption{Additional attention heatmap examples from the ChestXDet10 dataset.}
    \label{fig:appendix_cxd10_attn_maps}
\end{figure*}

We additionally visualize attention heatmaps to examine whether the pretrained model can perform reliable zero-shot grounding across diverse expressions and images. Fig.~\ref{fig:appendix_cxd10_attn_maps} presents additional examples from the ChestXDet10 dataset.

\vspace{-0.4cm}

\begin{figure*}[h!]
    \centering
    \centercaption
    \begin{minipage}{0.4\linewidth}
        \centering
        \includegraphics[width=\linewidth]{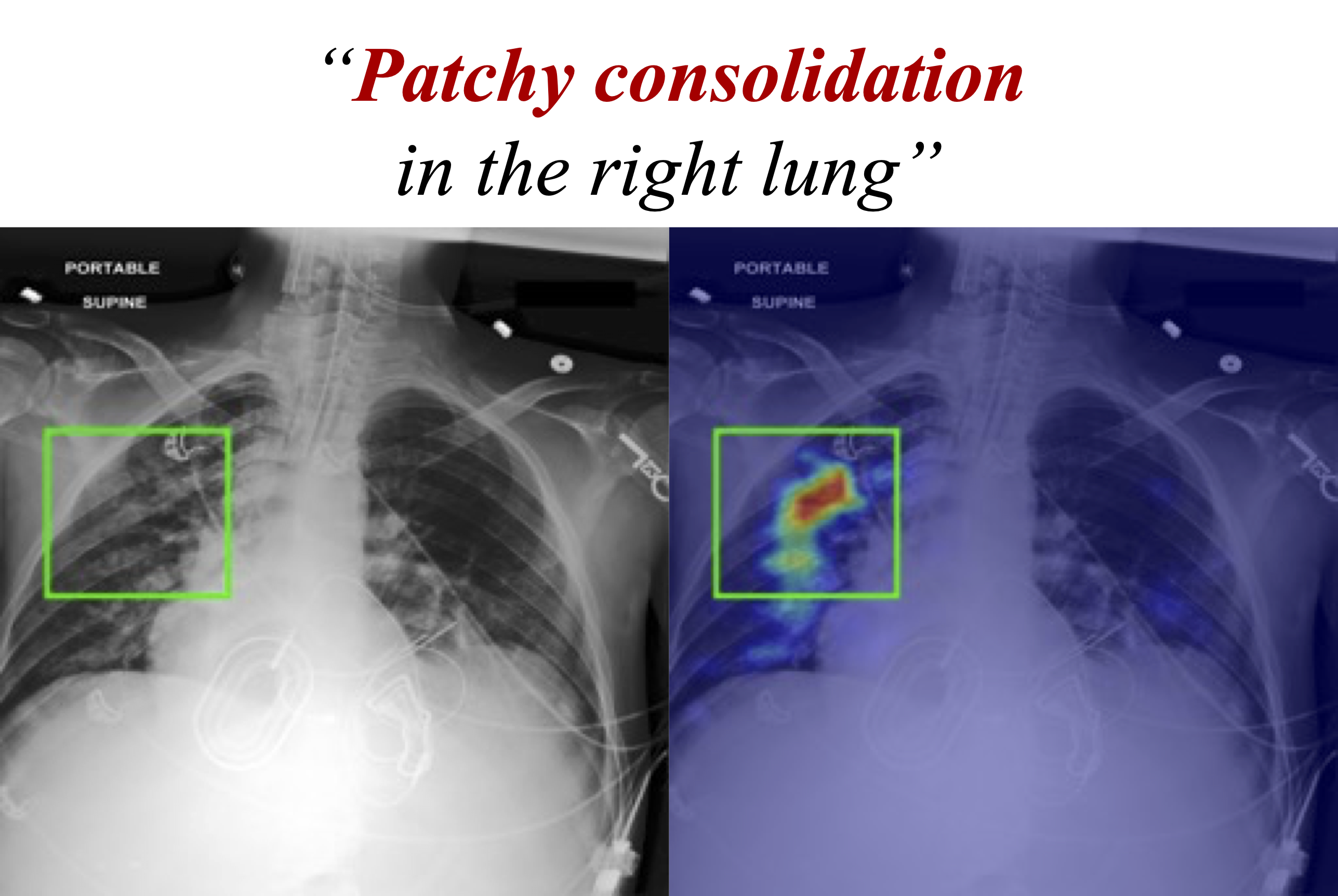}
    \end{minipage}
    \hspace{0.02\linewidth}% adjust as desired
    \begin{minipage}{0.4\linewidth}
        \centering
        \includegraphics[width=\linewidth]{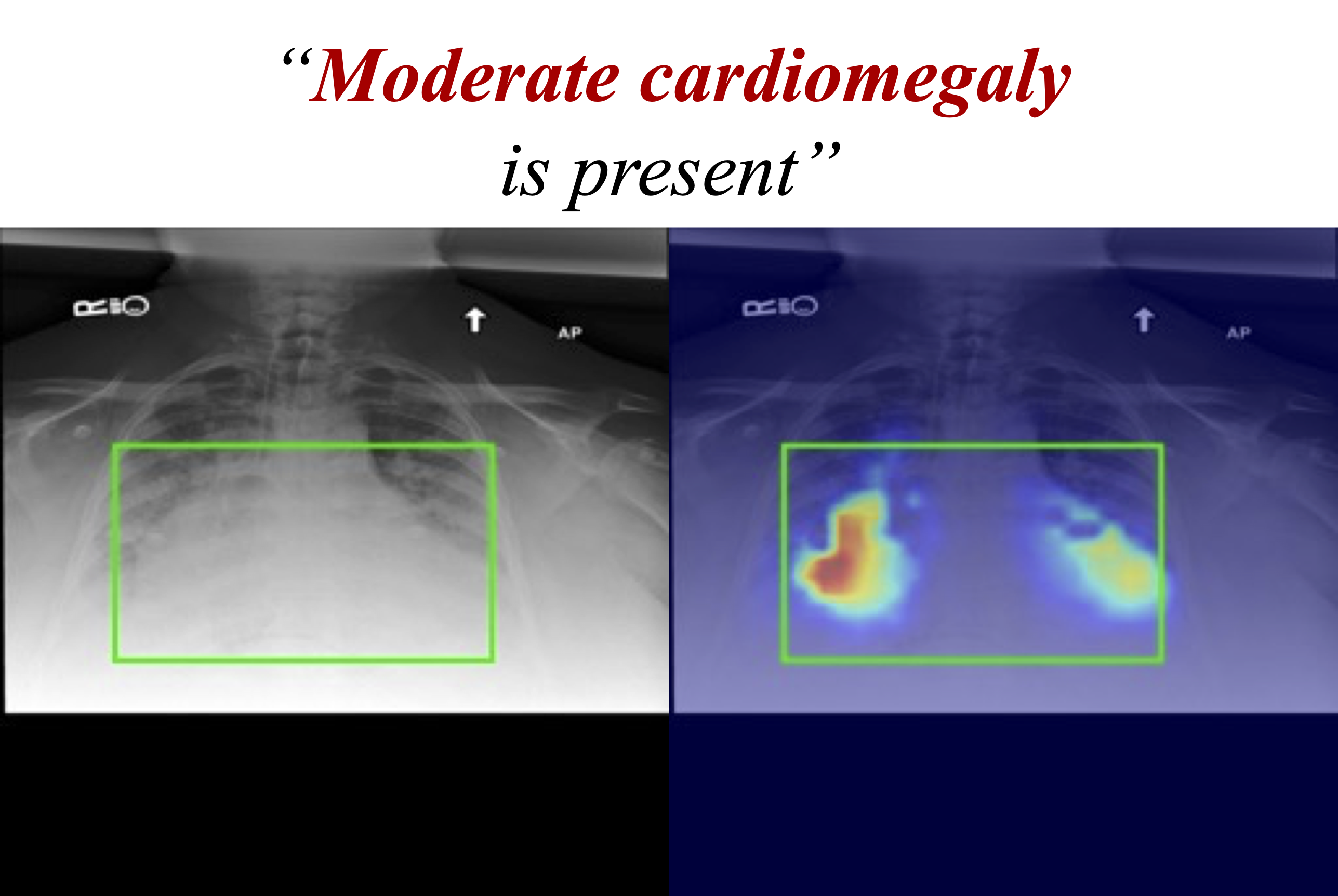}
    \end{minipage}

    \vspace{0.1cm}

    \begin{minipage}{0.4\linewidth}
        \centering
        \includegraphics[width=\linewidth]{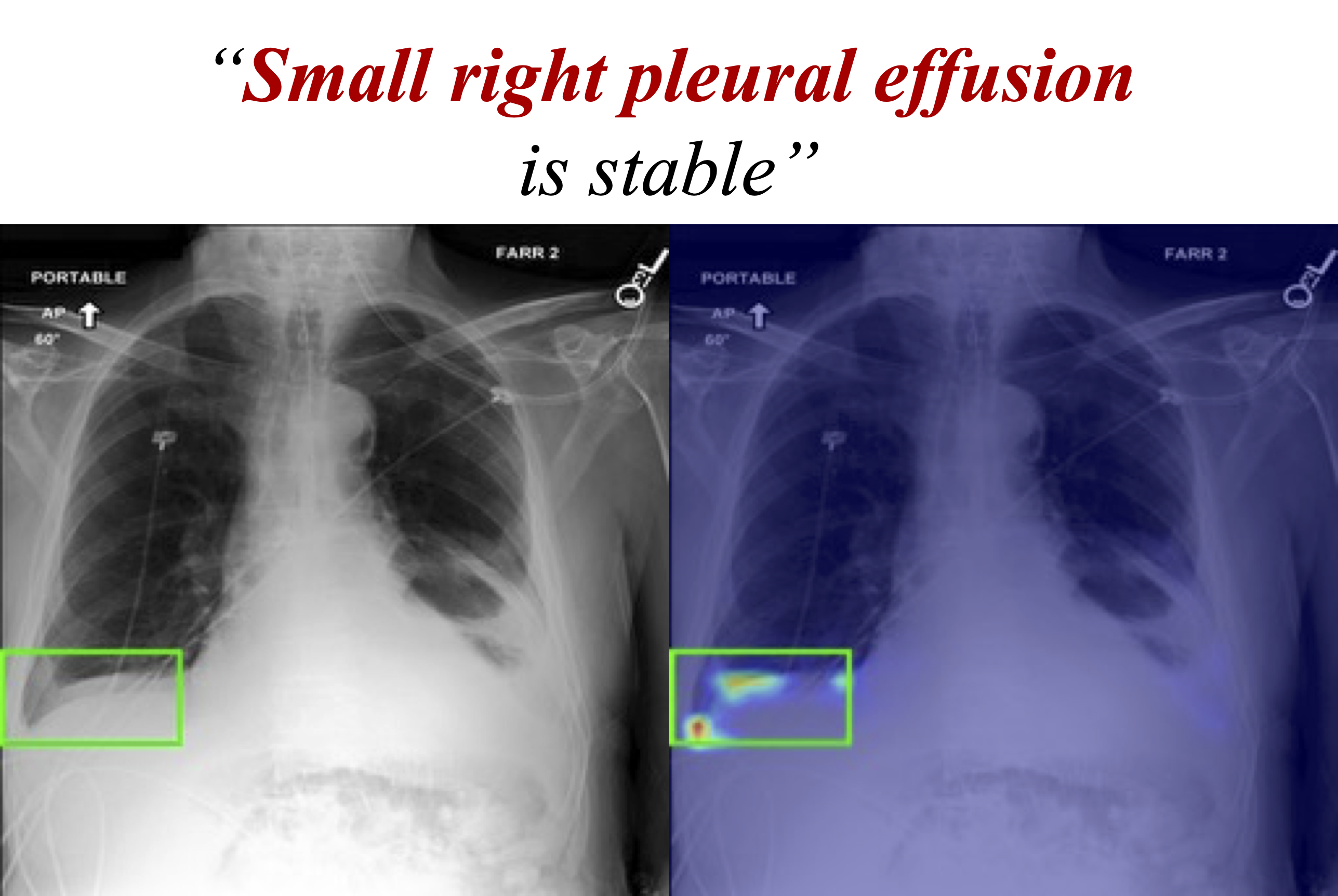}
    \end{minipage}
    \hspace{0.02\linewidth}% adjust as desired
    \begin{minipage}{0.4\linewidth}
        \centering
        \includegraphics[width=\linewidth]{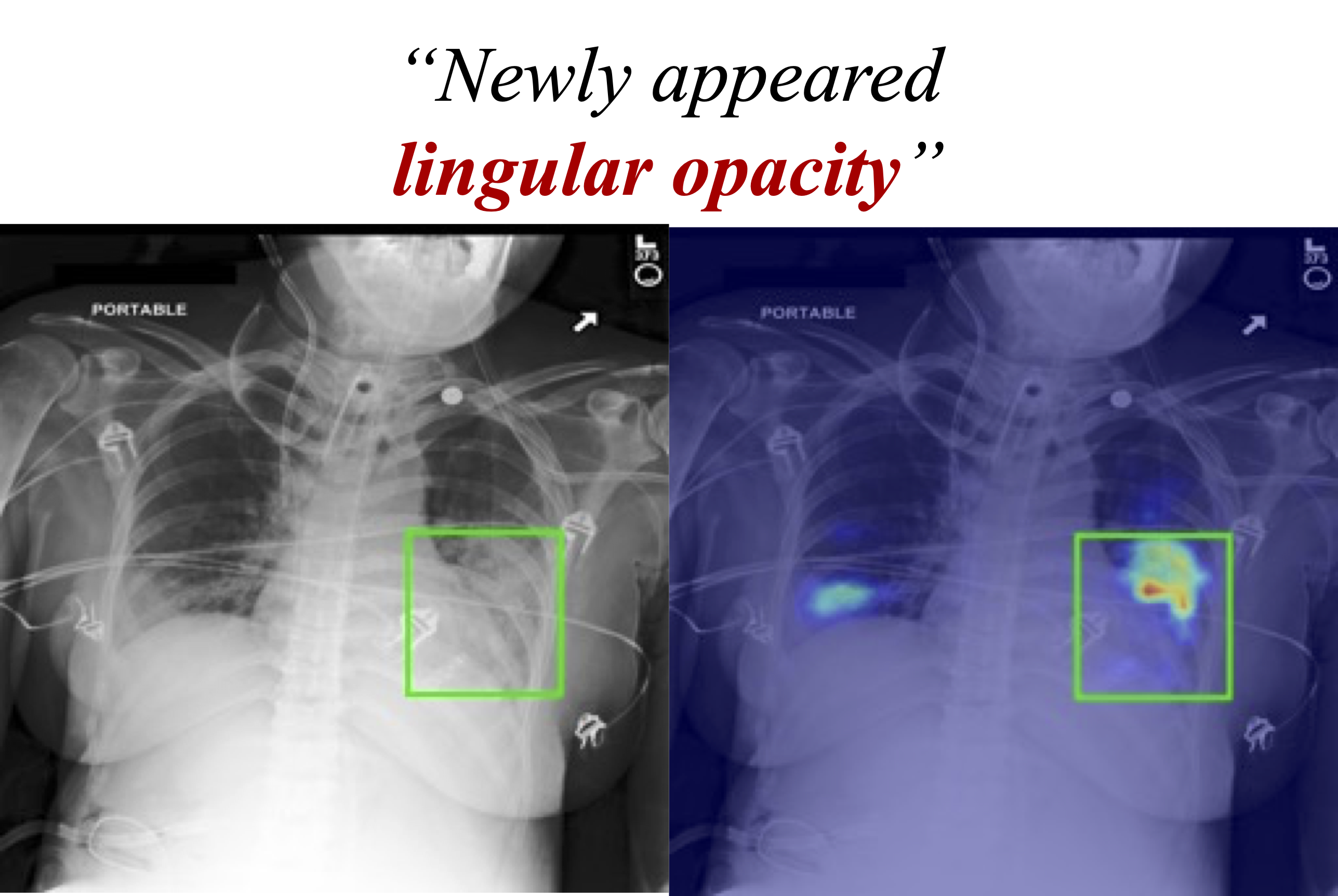}
    \end{minipage}

    \caption{Additional attention heatmap examples from the MS-CXR dataset.}
    \label{fig:appendix_ms_cxr_attn_maps}
\end{figure*}

\vspace{-0.3cm}
We also visualize attention maps for samples from the MS-CXR dataset in Fig.~\ref{fig:appendix_ms_cxr_attn_maps}. Compared to ChestXDet10, MS-CXR contains longer and more detailed text phrases paired with corresponding bounding boxes, requiring more fine-grained visual grounding. These results show that our proposed model can perform zero-shot visual grounding not only for short text prompts, but also for longer and more detailed clinical descriptions.

%% file: table/ablation_text_feature.tex
\begin{table*}[h]
\centering
\centercaption
\small
\caption{Ablation results of text feature conditioning in feature modulation units.}
\vspace{-0.2cm}
\resizebox{0.85\textwidth}{!}{%
\begin{tabular}{l|ccccc|cc}
\toprule
\multirow{2}{*}{}
& \multicolumn{5}{c|}{Classification}
& \multicolumn{2}{c}{Grounding} \\
\cmidrule(lr){2-6}
\cmidrule(lr){7-8}
& OpenI & CXR14 & PadChest & CXD10 & CheXPert
& CXD10 & MS-CXR \\
\midrule
w/o text feature & 0.8804 & 0.8358 & 0.8599 & \textbf{0.8523} & 0.9000 & 0.7298 & 0.9042 \\
w/ text feature & \textbf{0.8891} & \textbf{0.8389} & \textbf{0.8609} & 0.8433 & \textbf{0.9037} & \textbf{0.7315} & \textbf{0.9222} \\
\bottomrule
\end{tabular}%
}
\label{tab:text_feature_ablation}
\end{table*}

%% file: table/hyperparam_false_neg_w.tex
\begin{table*}[h]
\centering
\centercaption
\small
\caption{Effect of false negative loss coefficient.}
\vspace{-0.2cm}
\resizebox{0.75\textwidth}{!}{%
\begin{tabular}{l|ccccc|cc}
\toprule
\multirow{2}{*}{$\lambda_{fn}$}
& \multicolumn{5}{c|}{Classification}
& \multicolumn{2}{c}{Grounding} \\
\cmidrule(lr){2-6}
\cmidrule(lr){7-8}
& OpenI & CXR14 & PadChest & CXD10 & CheXPert
& CXD10 & MS-CXR \\
\midrule
0.01 & 0.8891 & 0.8389 & 0.8609 & 0.8433 & 0.9037 & 0.7315 & 0.9222 \\
0.03 & 0.8851 & 0.8381 & 0.8637 & 0.8490 & 0.9074 & 0.7018 & 0.8922 \\
0.05 & 0.8882 & 0.8377 & 0.8591 & 0.8510 & 0.9047 & 0.7209 & 0.9341 \\
\bottomrule
\end{tabular}%
}
\label{tab:hyperparam_fn_coeff}
\end{table*}

%% file: table/hyperparam_false_neg_k.tex
\begin{table*}[h]
\centering
\centercaption
\small
\caption{Effect of false negative patch sampling ratio, with $\lambda_{fn}=0.01$.}
\vspace{-0.2cm}
\resizebox{0.75\textwidth}{!}{%
\begin{tabular}{l|ccccc|cc}
\toprule
\multirow{2}{*}{$M$}
& \multicolumn{5}{c|}{Classification}
& \multicolumn{2}{c}{Grounding} \\
\cmidrule(lr){2-6}
\cmidrule(lr){7-8}
& OpenI & CXR14 & PadChest & CXD10 & CheXPert
& CXD10 & MS-CXR \\
\midrule
10\% & 0.8815 & 0.8350 & 0.8621 & 0.8457 & 0.9039 & 0.6998 & 0.9281 \\
20\% & 0.8891 & 0.8389 & 0.8609 & 0.8433 & 0.9037 & 0.7315 & 0.9222 \\
30\% & 0.8818 & 0.8335 & 0.8613 & 0.8489 & 0.8951 & 0.7168 & 0.9341 \\
% 50\% & 0.0000 & 0.0000 & 0.0000 & 0.0000 & 0.0000 & 0.0000 & 0.0000 \\
\bottomrule
\end{tabular}%
}
\label{tab:hyperparam_fn_ratio}
\end{table*}

%% file: paper.bbl
\begin{thebibliography}{33}
\providecommand{\natexlab}[1]{#1}
\providecommand{\url}[1]{\texttt{#1}}
\expandafter\ifx\csname urlstyle\endcsname\relax
  \providecommand{\doi}[1]{doi: #1}\else
  \providecommand{\doi}{doi: \begingroup \urlstyle{rm}\Url}\fi

\bibitem[Ba et~al.(2016)Ba, Kiros, and Hinton]{ba2016layer}
Jimmy~Lei Ba, Jamie~Ryan Kiros, and Geoffrey~E Hinton.
\newblock Layer normalization.
\newblock \emph{arXiv preprint arXiv:1607.06450}, 2016.

\bibitem[Bannur et~al.(2023)Bannur, Hyland, Liu, Perez-Garcia, Ilse, Castro, Boecking, Sharma, Bouzid, Thieme, et~al.]{bannur2023learning}
Shruthi Bannur, Stephanie Hyland, Qianchu Liu, Fernando Perez-Garcia, Maximilian Ilse, Daniel~C Castro, Benedikt Boecking, Harshita Sharma, Kenza Bouzid, Anja Thieme, et~al.
\newblock Learning to exploit temporal structure for biomedical vision-language processing.
\newblock In \emph{Proceedings of the IEEE/CVF conference on computer vision and pattern recognition}, pages 15016--15027, 2023.

\bibitem[Boecking et~al.(2022)Boecking, Usuyama, Bannur, Castro, Schwaighofer, Hyland, Wetscherek, Naumann, Nori, Alvarez-Valle, et~al.]{boecking2022making}
Benedikt Boecking, Naoto Usuyama, Shruthi Bannur, Daniel~C Castro, Anton Schwaighofer, Stephanie Hyland, Maria Wetscherek, Tristan Naumann, Aditya Nori, Javier Alvarez-Valle, et~al.
\newblock Making the most of text semantics to improve biomedical vision--language processing.
\newblock In \emph{European conference on computer vision}, pages 1--21. Springer, 2022.

\bibitem[Bustos et~al.(2020)Bustos, Pertusa, Salinas, and De~La Iglesia-Vaya]{bustos2020padchest}
Aurelia Bustos, Antonio Pertusa, Jose-Maria Salinas, and Maria De~La Iglesia-Vaya.
\newblock Padchest: A large chest x-ray image dataset with multi-label annotated reports.
\newblock \emph{Medical image analysis}, 66:\penalty0 101797, 2020.

\bibitem[Chen et~al.(2023)Chen, Zhou, Tran, Zhao, Wan, Ooi, Cheng, Thng, Xu, Liu, et~al.]{chen2023medical}
Zhihao Chen, Yang Zhou, Anh Tran, Junting Zhao, Liang Wan, Gideon Su~Kai Ooi, Lionel Tim-Ee Cheng, Choon~Hua Thng, Xinxing Xu, Yong Liu, et~al.
\newblock Medical phrase grounding with region-phrase context contrastive alignment.
\newblock In \emph{International Conference on Medical Image Computing and Computer-Assisted Intervention}, pages 371--381. Springer, 2023.

\bibitem[Cheng et~al.(2023)Cheng, Lin, Lyu, Huang, Luo, and Tang]{cheng2023prior}
Pujin Cheng, Li~Lin, Junyan Lyu, Yijin Huang, Wenhan Luo, and Xiaoying Tang.
\newblock Prior: Prototype representation joint learning from medical images and reports.
\newblock In \emph{Proceedings of the IEEE/CVF international conference on computer vision}, pages 21361--21371, 2023.

\bibitem[Demner-Fushman et~al.(2016)Demner-Fushman, Kohli, Rosenman, Shooshan, Rodriguez, Antani, Thoma, and McDonald]{demner2016preparing}
Dina Demner-Fushman, Marc~D Kohli, Marc~B Rosenman, Sonya~E Shooshan, Laritza Rodriguez, Sameer Antani, George~R Thoma, and Clement~J McDonald.
\newblock Preparing a collection of radiology examinations for distribution and retrieval.
\newblock \emph{Journal of the American Medical Informatics Association}, 23\penalty0 (2):\penalty0 304--310, 2016.

\bibitem[Dosovitskiy et~al.(2021)Dosovitskiy, Beyer, Kolesnikov, Weissenborn, Zhai, Unterthiner, Dehghani, Minderer, Heigold, Gelly, Uszkoreit, and Houlsby]{dosovitskiy2020vit}
Alexey Dosovitskiy, Lucas Beyer, Alexander Kolesnikov, Dirk Weissenborn, Xiaohua Zhai, Thomas Unterthiner, Mostafa Dehghani, Matthias Minderer, Georg Heigold, Sylvain Gelly, Jakob Uszkoreit, and Neil Houlsby.
\newblock An image is worth 16x16 words: Transformers for image recognition at scale.
\newblock \emph{ICLR}, 2021.

\bibitem[Huang et~al.(2021)Huang, Shen, Lungren, and Yeung]{huang2021gloria}
Shih-Cheng Huang, Liyue Shen, Matthew~P Lungren, and Serena Yeung.
\newblock Gloria: A multimodal global-local representation learning framework for label-efficient medical image recognition.
\newblock In \emph{Proceedings of the IEEE/CVF international conference on computer vision}, pages 3942--3951, 2021.

\bibitem[Irvin et~al.(2019)Irvin, Rajpurkar, Ko, Yu, Ciurea-Ilcus, Chute, Marklund, Haghgoo, Ball, Shpanskaya, et~al.]{irvin2019chexpert}
Jeremy Irvin, Pranav Rajpurkar, Michael Ko, Yifan Yu, Silviana Ciurea-Ilcus, Chris Chute, Henrik Marklund, Behzad Haghgoo, Robyn Ball, Katie Shpanskaya, et~al.
\newblock Chexpert: A large chest radiograph dataset with uncertainty labels and expert comparison.
\newblock In \emph{Proceedings of the AAAI Conference on Artificial Intelligence}, volume~33, pages 590--597, 2019.

\bibitem[Johnson et~al.(2024)Johnson, Pollard, Mark, Berkowitz, and Horng]{mimic_cxr_physionet}
Alistair Johnson, Tom Pollard, Roger Mark, Seth Berkowitz, and Steven Horng.
\newblock {MIMIC-CXR Database}.
\newblock \emph{{PhysioNet}}, July 2024.
\newblock \doi{10.13026/4jqj-jw95}.
\newblock URL \url{https://doi.org/10.13026/4jqj-jw95}.
\newblock Version 2.1.0.

\bibitem[Johnson et~al.(2019)Johnson, Pollard, Berkowitz, Greenbaum, Lungren, Deng, Mark, and Horng]{johnson2019mimic}
Alistair~EW Johnson, Tom~J Pollard, Seth~J Berkowitz, Nathaniel~R Greenbaum, Matthew~P Lungren, Chih-ying Deng, Roger~G Mark, and Steven Horng.
\newblock Mimic-cxr, a de-identified publicly available database of chest radiographs with free-text reports.
\newblock \emph{Scientific data}, 6\penalty0 (1):\penalty0 317, 2019.

\bibitem[Lai et~al.(2024)Lai, Yao, Jiang, Wang, He, Tao, and Zhou]{lai2024carzero}
Haoran Lai, Qingsong Yao, Zihang Jiang, Rongsheng Wang, Zhiyang He, Xiaodong Tao, and S~Kevin Zhou.
\newblock Carzero: Cross-attention alignment for radiology zero-shot classification.
\newblock In \emph{Proceedings of the IEEE/CVF Conference on Computer Vision and Pattern Recognition}, pages 11137--11146, 2024.

\bibitem[Lee et~al.(2022)Lee, Kim, Shon, Kim, Kim, Lee, and Kim]{lee2022uniclip}
Janghyeon Lee, Jongsuk Kim, Hyounguk Shon, Bumsoo Kim, Seung~Hwan Kim, Honglak Lee, and Junmo Kim.
\newblock Uniclip: Unified framework for contrastive language-image pre-training.
\newblock \emph{Advances in Neural Information Processing Systems}, 35:\penalty0 1008--1019, 2022.

\bibitem[Li et~al.(2024)Li, Yang, Ren, Nie, Gao, Tan, and Li]{li2024mlip}
Zhe Li, Laurence~T Yang, Bocheng Ren, Xin Nie, Zhangyang Gao, Cheng Tan, and Stan~Z Li.
\newblock Mlip: Enhancing medical visual representation with divergence encoder and knowledge-guided contrastive learning.
\newblock In \emph{Proceedings of the IEEE/CVF Conference on Computer Vision and Pattern Recognition}, pages 11704--11714, 2024.

\bibitem[Lian et~al.(2026)Lian, Zhou, Wong, and Qin]{lian2026concept}
Chenyu Lian, Hong-Yu Zhou, Chun-Ka Wong, and Jing Qin.
\newblock Concept-guided noisy negative suppression for zero-shot classification and grounding of chest x-ray findings.
\newblock \emph{arXiv preprint arXiv:2605.19374}, 2026.

\bibitem[Liu et~al.(2024)Liu, Lu, and Wang]{liu2024towards}
Bo~Liu, Zexin Lu, and Yan Wang.
\newblock Towards medical vision-language contrastive pre-training via study-oriented semantic exploration.
\newblock In \emph{Proceedings of the 32nd ACM International Conference on Multimedia}, pages 4861--4870, 2024.

\bibitem[Liu et~al.(2020)Liu, Lian, and Yu]{liu2020chestx}
Jingyu Liu, Jie Lian, and Yizhou Yu.
\newblock Chestx-det10: chest x-ray dataset on detection of thoracic abnormalities.
\newblock \emph{arXiv preprint arXiv:2006.10550}, 2020.

\bibitem[Oord et~al.(2018)Oord, Li, and Vinyals]{oord2018representation}
Aaron van~den Oord, Yazhe Li, and Oriol Vinyals.
\newblock Representation learning with contrastive predictive coding.
\newblock \emph{arXiv preprint arXiv:1807.03748}, 2018.

\bibitem[Park et~al.(2025)Park, Kim, Yoon, and Choi]{park2025radzero}
Jonggwon Park, Soobum Kim, Byungmu Yoon, and Kyoyun Choi.
\newblock Radzero: Similarity-based cross-attention for explainable vision-language alignment in radiology with zero-shot multi-task capability.
\newblock \emph{arXiv e-prints}, pages arXiv--2504, 2025.

\bibitem[Park et~al.(2026)Park, Lee, Park, Yun, Kim, Jeong, Kang, Yoon, and Choi]{park2026glint}
Jonggwon Park, Seongeun Lee, Junhyun Park, Hannah Yun, Hyunwoong Kim, Sohyun Jeong, Hyewon Kang, Byungmu Yoon, and Kyoyun Choi.
\newblock Glint: Sparsely gated vision-language alignment for fine-grained radiology representations.
\newblock \emph{arXiv preprint arXiv:2606.03180}, 2026.

\bibitem[Perez et~al.(2018)Perez, Strub, De~Vries, Dumoulin, and Courville]{perez2018film}
Ethan Perez, Florian Strub, Harm De~Vries, Vincent Dumoulin, and Aaron Courville.
\newblock Film: Visual reasoning with a general conditioning layer.
\newblock In \emph{Proceedings of the AAAI conference on artificial intelligence}, 2018.

\bibitem[P{\'e}rez-Garc{\'\i}a et~al.(2025)P{\'e}rez-Garc{\'\i}a, Sharma, Bond-Taylor, Bouzid, Salvatelli, Ilse, Bannur, Castro, Schwaighofer, Lungren, et~al.]{perez2025exploring}
Fernando P{\'e}rez-Garc{\'\i}a, Harshita Sharma, Sam Bond-Taylor, Kenza Bouzid, Valentina Salvatelli, Maximilian Ilse, Shruthi Bannur, Daniel~C Castro, Anton Schwaighofer, Matthew~P Lungren, et~al.
\newblock Exploring scalable medical image encoders beyond text supervision.
\newblock \emph{Nature Machine Intelligence}, 7\penalty0 (1):\penalty0 119--130, 2025.

\bibitem[Radford et~al.(2021)Radford, Kim, Hallacy, Ramesh, Goh, Agarwal, Sastry, Askell, Mishkin, Clark, et~al.]{radford2021learning}
Alec Radford, Jong~Wook Kim, Chris Hallacy, Aditya Ramesh, Gabriel Goh, Sandhini Agarwal, Girish Sastry, Amanda Askell, Pamela Mishkin, Jack Clark, et~al.
\newblock Learning transferable visual models from natural language supervision.
\newblock In \emph{International conference on machine learning}, pages 8748--8763. PmLR, 2021.

\bibitem[Reimers and Gurevych(2019)]{reimers2019sentence}
Nils Reimers and Iryna Gurevych.
\newblock Sentence-bert: Sentence embeddings using siamese bert-networks.
\newblock In \emph{Proceedings of the 2019 conference on empirical methods in natural language processing and the 9th international joint conference on natural language processing (EMNLP-IJCNLP)}, pages 3982--3992, 2019.

\bibitem[Song et~al.(2020)Song, Tan, Qin, Lu, and Liu]{song2020mpnet}
Kaitao Song, Xu~Tan, Tao Qin, Jianfeng Lu, and Tie-Yan Liu.
\newblock Mpnet: Masked and permuted pre-training for language understanding.
\newblock \emph{Advances in neural information processing systems}, 33:\penalty0 16857--16867, 2020.

\bibitem[Wang et~al.(2022)Wang, Zhou, Wang, Vardhanabhuti, and Yu]{wang2022multi}
Fuying Wang, Yuyin Zhou, Shujun Wang, Varut Vardhanabhuti, and Lequan Yu.
\newblock Multi-granularity cross-modal alignment for generalized medical visual representation learning.
\newblock \emph{Advances in neural information processing systems}, 35:\penalty0 33536--33549, 2022.

\bibitem[Wang et~al.(2017)Wang, Peng, Lu, Lu, Bagheri, and Summers]{wang2017chestx}
Xiaosong Wang, Yifan Peng, Le~Lu, Zhiyong Lu, Mohammadhadi Bagheri, and Ronald~M Summers.
\newblock Chestx-ray8: Hospital-scale chest x-ray database and benchmarks on weakly-supervised classification and localization of common thorax diseases.
\newblock In \emph{Proceedings of the IEEE conference on computer vision and pattern recognition}, pages 2097--2106, 2017.

\bibitem[Wu et~al.(2023)Wu, Zhang, Zhang, Wang, and Xie]{wu2023medklip}
Chaoyi Wu, Xiaoman Zhang, Ya~Zhang, Yanfeng Wang, and Weidi Xie.
\newblock Medklip: Medical knowledge enhanced language-image pre-training for x-ray diagnosis.
\newblock In \emph{Proceedings of the IEEE/CVF international conference on computer vision}, pages 21372--21383, 2023.

\bibitem[Zhang et~al.(2018)Zhang, Bargal, Lin, Brandt, Shen, and Sclaroff]{zhang2018top}
Jianming Zhang, Sarah~Adel Bargal, Zhe Lin, Jonathan Brandt, Xiaohui Shen, and Stan Sclaroff.
\newblock Top-down neural attention by excitation backprop.
\newblock \emph{International Journal of Computer Vision}, 126\penalty0 (10):\penalty0 1084--1102, 2018.

\bibitem[Zhang et~al.(2023)Zhang, Wu, Zhang, Xie, and Wang]{zhang2023knowledge}
Xiaoman Zhang, Chaoyi Wu, Ya~Zhang, Weidi Xie, and Yanfeng Wang.
\newblock Knowledge-enhanced visual-language pre-training on chest radiology images.
\newblock \emph{Nature Communications}, 14\penalty0 (1):\penalty0 4542, 2023.

\bibitem[Zhang et~al.(2022)Zhang, Jiang, Miura, Manning, and Langlotz]{zhang2022contrastive}
Yuhao Zhang, Hang Jiang, Yasuhide Miura, Christopher~D Manning, and Curtis~P Langlotz.
\newblock Contrastive learning of medical visual representations from paired images and text.
\newblock In \emph{Machine learning for healthcare conference}, pages 2--25. PMLR, 2022.

\bibitem[Zhang et~al.(2025)Zhang, Yu, Chen, Yang, and Yeo]{zhang2025medunifier}
Ziyang Zhang, Yang Yu, Yucheng Chen, Xulei Yang, and Si~Yong Yeo.
\newblock Medunifier: Unifying vision-and-language pre-training on medical data with vision generation task using discrete visual representations.
\newblock In \emph{Proceedings of the IEEE/CVF Conference on Computer Vision and Pattern Recognition}, pages 29744--29755, 2025.

\end{thebibliography}
